\documentclass[]{article}
\usepackage[a4paper, portrait, margin=1in]{geometry}

\usepackage{authblk}
\usepackage{amsmath,amsfonts,amssymb}
\usepackage{graphicx}
\usepackage{tocloft}

\usepackage{amsmath}
\usepackage{booktabs}
\usepackage{tabularx}
\usepackage{wrapfig}
\usepackage{graphicx}
\usepackage{subfig}
\usepackage{booktabs}
\usepackage{afterpage}
\usepackage{rotating}
\usepackage{forest}
\usepackage[]{siunitx}
\usepackage{url}
\usepackage[space]{grffile}
\usepackage[inline]{enumitem}
\usepackage{lscape}

\usepackage{lineno,hyperref}
\modulolinenumbers[5]

\providecommand{\keywords}[1]{\textbf{\textit{Index terms---}} #1}

\begin{document} 
\title{Machine learning based hyperspectral image analysis: A survey}
\author[1,*]{Utsav B. Gewali}
\author[2]{Sildomar T. Monteiro}
\author[1,3]{Eli Saber}
\affil[1]{Chester F. Carlson Center for Imaging Science, Rochester Institute of Technology, Rochester, NY}
\affil[2]{The Boeing Company, Huntsville, AL}
\affil[3]{Department of Electrical \& Microelectronic Engineering, Rochester Institute of Technology, Rochester, NY}
\affil[*]{\tt{ubg9540@rit.edu}}

\date{}
\maketitle

\begin{abstract}
Hyperspectral sensors enable the study of the chemical and physical properties of scene materials remotely for the purpose of identification, detection, chemical composition analysis, and physical parameter estimation of objects in the environment. Hence, hyperspectral images captured from earth observing satellites and aircraft have been increasingly important in agriculture, environmental monitoring, urban planning, mining, and defense. Machine learning algorithms due to their outstanding predictive power have become a key tool for modern hyperspectral image analysis. Therefore, a solid understanding of machine learning techniques have become essential for remote sensing researchers and practitioners. This paper surveys and compares recent machine learning-based hyperspectral image analysis methods published in literature. We organize the methods by the image analysis task and by the type of machine learning algorithm, and present a two-way mapping between the image analysis tasks and the types of machine learning algorithms that can be applied to them. The paper is comprehensive in coverage of both hyperspectral image analysis tasks and machine learning algorithms. The image analysis tasks considered are land cover classification, target/anomaly detection, unmixing, and physical/chemical parameter estimation. The machine learning algorithms covered are Gaussian models, linear regression, logistic regression, support vector machines, Gaussian mixture model, latent linear models, sparse linear models, Gaussian mixture models, ensemble learning, directed graphical models, undirected graphical models, clustering, Gaussian processes, Dirichlet processes, and deep learning. We also discuss the open challenges in the field of hyperspectral image analysis and explore possible future directions.   
\end{abstract}


\section{Introduction}
With applications to fields such as agriculture~\cite{dale2013}, ecology~\cite{wang2010}, mining~\cite{van2012}, forestry~\cite{ghiyamat2010}, urban planning~\cite{wentz2014}, defense~\cite{yuen2010}, and space exploration~\cite{pilorget2014}, hyperspectral imaging is a powerful remote sensing modality to study the chemical and physical properties of scene materials. Hyperspectral imaging, also known as imaging spectroscopy, captures the reflected or emitted electromagnetic energy from a scene over hundreds of narrow, contiguous spectral bands, from visible to infrared wavelengths~\cite{eismann2012}. Each pixel in a hyperspectral image is composed of a vector of hundreds of elements measuring the reflected or emitted energy as a function of wavelength, known as the spectrum. The spectrum captures the information about the material's chemical and physical properties because the interaction between light at different wavelengths and the material is governed by material's atomic and molecular structure. Hence, hyperspectral sensors mounted on aircrafts and satellites can collect images over a large geographical area which can be automatically analyzed by algorithms to map the properties of materials in the scene.

\begin{figure}[!h]
\centering
\includegraphics[width=0.5\linewidth]{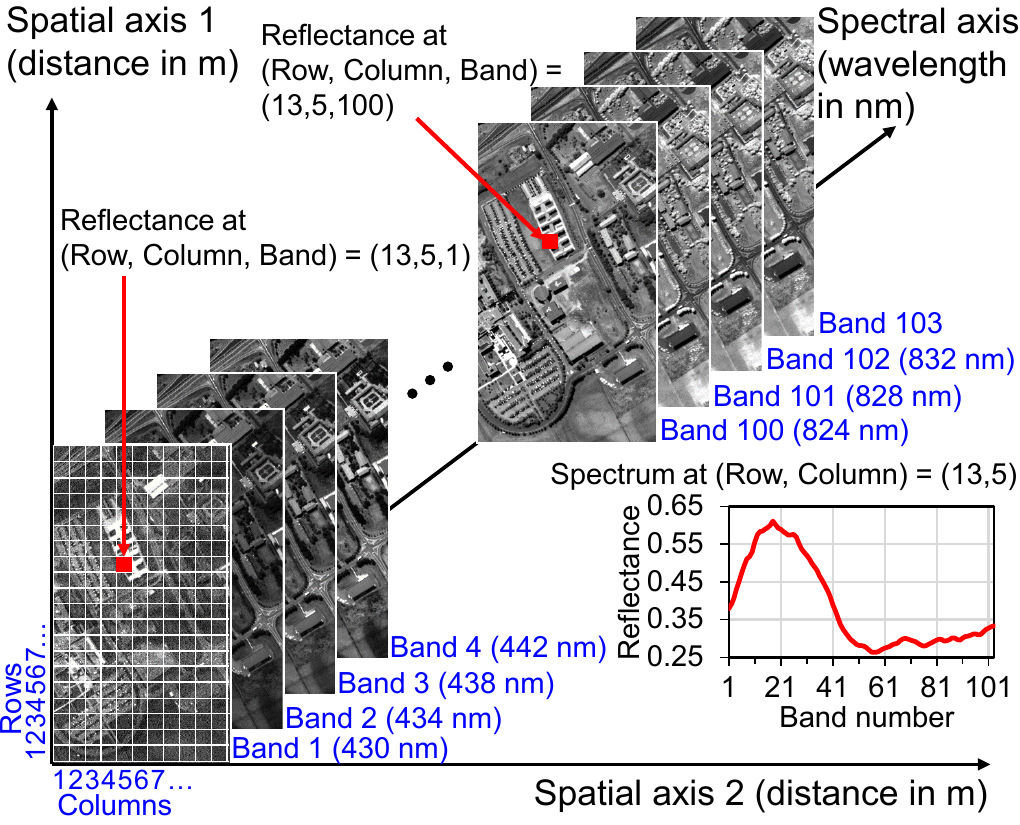}
\caption{Hyperspectral image as a data cube.}
\label{fig:datacube}
\end{figure}

\begin{figure}[!h]
\centering
\includegraphics[width=0.4\linewidth]{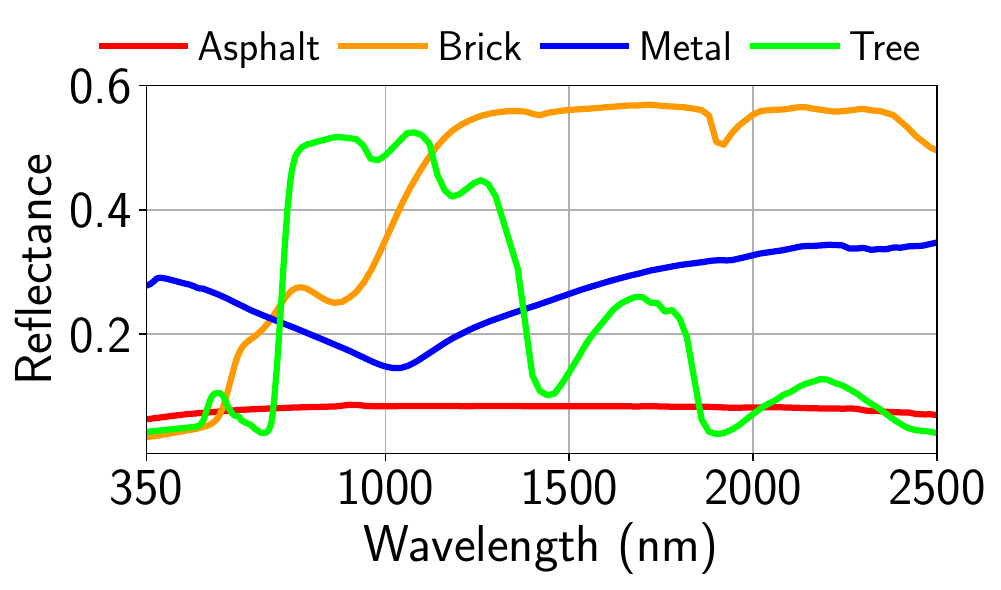}
\caption{Sample spectra of different materials from a spectral library~\cite{baldridge2009aster}.}
\label{fig:spectrallib}
\end{figure}

\begin{figure}[!h]
\centering
\includegraphics[width=0.5\linewidth]{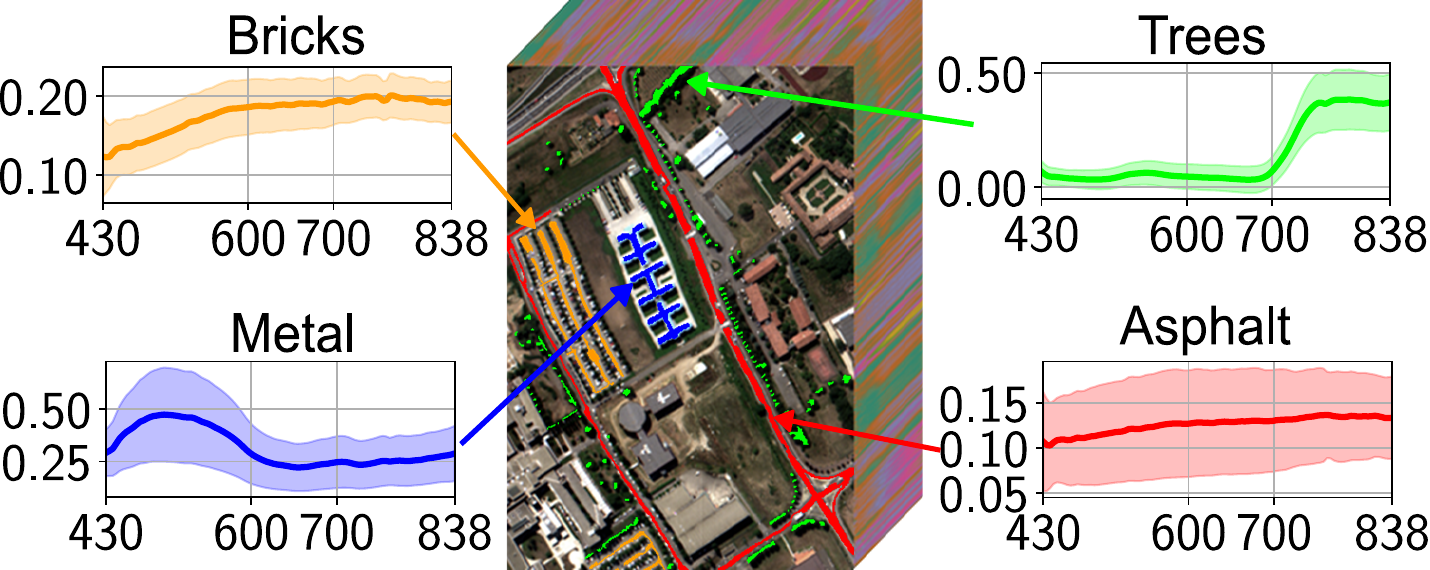}
\caption{Spectral variability within an image. The x- and y-axes of the plots are wavelength in \SI{}{\nano\meter} and reflectance respectively.}
\label{fig:spectralvar}
\end{figure}

In this survey, we primarily discuss reflective hyperspectral remote sensing, in which spectra are measured over the entirety or over a subset of the visible, near infrared, and shortwave infrared wavelengths  (\SI{350}{\nano\metre} to \SI{2500}{\nano\metre}). The radiance reaching the sensor in this spectral range is dominated by the solar energy reflected from the objects in the scene and it is proportional to the directional reflectance of the objects' surface materials. The spectra are measured in terms of either observed radiance or surface reflectance. Reflectance as a unit of spectrum is generally preferred because it is an intrinsic material property independent of illumination. Surface reflectance is estimated from the observed radiance by using atmospheric compensation techniques~\cite{gao2009}. Fig.~\ref{fig:datacube} shows a hyperspectral image as a three dimensional data cube, with spatial axes sampled into pixels and the reflectance spectrum at each pixel sampled into bands. Each element in the data cube represents a reflectance value averaged over the area covered by a particular pixel (indexed by row and column numbers) and integrated over a given band of wavelengths (indexed by band number). The shape of reflectance spectrum, sometimes called spectral signature, is generally unique to a material and can be used to identify and study materials~\cite{price1994unique}. For illustration, Fig.~\ref{fig:spectrallib} shows the differences in the spectral signatures of four different materials. However, measured spectra always exhibit variabilities~\cite{zare2014endmember} which make the data analysis difficult. Spectral variabilities are not only observed between different images but also seen within a single image.  Fig.~\ref{fig:spectralvar} plots the empirical mean with 95\% confidence interval of spectra belonging to same ground covers in an image to demonstrate spectral variabilities in hyperspectral data. The image used in Figs.~\ref{fig:datacube} and \ref{fig:spectralvar} was obtained from \cite{pavia_university_cite}.


The variabilities in spectral signatures can be attributed to extrinsic or intrinsic factors. The extrinsic factors include differences in atmosphere~\cite{gao2009}, surrounding environment~\cite{gao2000optical}, illumination~\cite{wendel2017illumination}, sun-sensor geometry~\cite{galvao2004sun}, sensor~\cite{van2006multi} and any other factors which are not related to the properties of the material. The intrinsic factors are the ones related to the material itself, such as finer classification within the same material class~\cite{cochrane2000using}, or samples having different concentrations of constituent chemicals or different physical properties~\cite{asner1998biophysical}. A significant challenge for a successful image analysis algorithm is to be able to extract the desired information (signal) from intrinsic spectral variations while ignoring the extrinsic variations and intrinsic variation caused by unrelated factors. 

Another major challenge for hyperspectral image analysis algorithms is high dimensionality~\cite{landgrebe2002hyperspectral}. The dimensionality of spectra is equal to the total number of bands, with each band representing a dimension, and is large ranging in hundreds. When the number of dimensions is linearly increased, the volume of the feature space increases exponentially and hence enormous amount of data is required for modeling in this space~\cite{bishop2006}. However, due to difficulties in collection and costs associated with the analysis of material's chemical and physical properties, ground truth data is very scarce in hyperspectral datasets. This unfortunate combination of high dimensionality and limited ground truth data leads models to overfit and have low generalization performance. This problem has been referred in literature as the curse of dimensionality or the Hughes phenomenon. The classical approach to mitigate this problem is dimensionality reduction~\cite{hasanlou2012}, which is performed via feature extraction, i.e., transform the spectra to a lower dimensional representation, or band selection~\cite{bajcsy2004}, i.e., select only a subset of most significant bands for analysis. Feature extraction for dimensionality reduction is based on the hypothesis that hyperspectral bands oversample gradually-varying reflectance spectrum at most wavelengths, so there should be a more succinct representation of the spectral data. Similarly, band selection is based on the hypothesis that the effects of differences in material properties are only manifested in few bands, also called spectral features, so entire spectrum is unnecessary for analysis. In recent years, it has been popular to utilize spatial information along with spectral data during analysis to combat the problem of high dimensionality~\cite{fauvel2013,shi2014}. Neighboring pixels in high-resolution hyperspectral images are highly interdependent because most of the land covers are much bigger than the size of the pixel and presence of a material in one part of the image controls the likelihood of another material being present in another part of the image. Spatial-spectral methods model image analysis tasks as joint estimation over groups of interdependent pixels whose properties are constrained with one another rather than modeling as independent estimation over individual pixels, thus requiring less ground truth data for same level of accuracy~\cite{gewali2018tutorial}. 

Machine learning and pattern recognition based methods have been very successful for hyperspectral image analysis tasks, as they are able to automatically learn the relationship between the reflectance spectrum and the desired information while being robust against the noise and uncertainties in spectral and ground truth measurements. Studies have shown that machine learning-based methods can outperform traditional methods, such as spectral matching, manually designed normalized indices and physics-based modeling, e.g., Refs.~\cite{verrelst2012,schneider2010gaussian,darvishzadeh2011mapping,preidl2011comparison}. However, the bigger advantage of using machine learning-based methods is their flexibility. For example, a physics-based model, such as PROSPECT~\cite{feret2008prospect}, which was designed to model the variations in vegetation spectra due to six biophysical and biochemical parameters can only be used to study the distribution of those parameters in the image but not of any other parameter. Same is the case for a normalized index design for a particular physical/chemical parameter. Hence, these approaches cannot be easily used when there are no pre-existing models available for the physical/chemical property of interest. However, the same machine learning method, in theory, could be applied to study any physical/chemical parameter should enough ground truth data is available for training. Similarly, manually design material classifiers, such as USGS's tetracoder~\cite{clark2003imaging}, are not flexible enough to incorporate new material classes easily. Machine learning approaches can also generally better handle spectral and ground truth variability and noise compared to classical methods, such as spectral matching (which in machine learning terms is essentially nearest neighbor search with cosine distance metric)~\cite{schneider2010gaussian}. Techniques to tackle high dimensionality, such as dimensional reduction, band selection, and spatial-spectral predictions, can be easily incorporated with many classes of machine learning algorithms, e.g., Refs.~\cite{shaw2002_llm,cheng2006,liang2016_deep}. Additionally, Bayesian methods are a large class of machine learning algorithms which are designed to handle uncertainties and are good for modeling high-dimensional data~\cite{bishop2006}. Many studies~\cite{verrelst2013, eches2010, altmann2014_UGM,valls2016} have shown them to be well-suited for hyperspectral datasets.

\begin{figure}[!h]
\centering
\includegraphics[width=0.5\linewidth]{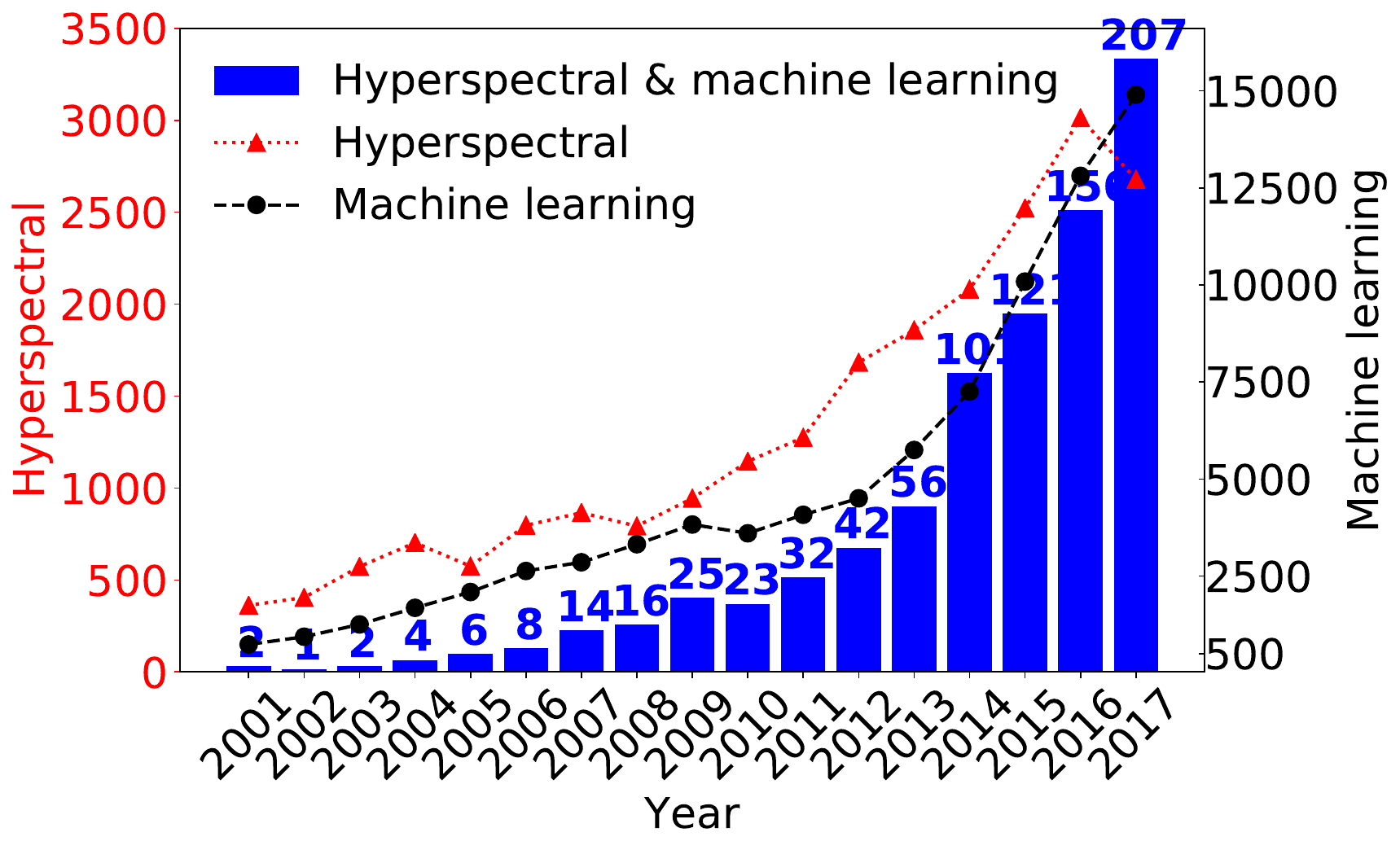}
\caption{Number of publications over the years with topics related to hyperspectral data and machine learning. Statistics obtained from Clarivate Analytics' Web of Science~\cite{webofscience}.}
\label{fig:pub_years}
\end{figure}

The remote sensing community has shown a great deal of interest in machine learning recently. Many journals have published special issues on machine learning for remote sensing~\cite{tuia2014_sp,chi2015_sp,alavi2016_sp,camps2016_sp}, numerous articles have been published on the topic of rise of machine learning in remotes sensing~\cite{valls2009_intro,lary2016_intro}, and all of the winning methods of the recent annual remote sensing GRSS data fusion competition~\cite{debes2014_df,liao2015_df,moser2015_df,tuia2016_df} and the top performing methods on ISPRS benchmark tests~\cite{ISPRSBenchmarkServer} have been based on machine learning. The hyperspectral remote sensing community has been equally active in this field and produced a great number of new methods. This can be see in Fig.~\ref{fig:pub_years} which shows how the total number of articles published each year whose topics deal with both hyperspectral data and machine learning has grown exponentially over the years. For comparison, we have also included two plots showing the total number of publications per year related to any topics in hyperspectral imaging (sensors, algorithms, and case studies) and any topics in machine learning (theory and applications), respectively. They have been increasing exponentially as well.  

This survey paper aims to provide a broad coverage of both the hyperspectral image analysis tasks and the machine learning algorithms, unlike previous surveys and tutorials which have either focused on a task~\cite{bioucas2012,lu2007,manolakis2002,valls2014,cheng2016survey} or a particular machine learning algorithm~\cite{valls2016,zhang2016,mountrakis2011,belgiu2016random,gewali2018tutorial}. All of the methods reviewed in the paper were published in peer-reviewed journals. The surveyed methods are able to analyze both radiance and reflectance images, unless otherwise stated. The hyperspectral data analysis tasks are categorized as land cover classification~\cite{valls2014}, target/anomaly detection~\cite{nasrabadi2014}, unmixing~\cite{bioucas2012}and physical/chemical parameter estimation~\cite{treitz1999}. We do not discuss multi-temporal/change analysis~\cite{hussain2013,somers2012hyperspectral}, preprocessing steps, such as, dimensionality reduction~\cite{khodr2011dimensionality} and feature extraction~\cite{lunga2014manifold}, or image processing tasks, such as, inpainting~\cite{chen2012inpainting}, denoising~\cite{yuan2012hyperspectral}, pansharpening/super-resolution~\cite{loncan2015hyperspectral}, and compression~\cite{christophe2011hyperspectral}, unless they are part of the image analysis method being discussed. The machine learning algorithms covered are Gaussian models~\cite{tong2012}, linear regression~\cite{montgomery2015}, logistic regression~\cite{menard2002}, support vector machines~\cite{scholkopf2002}, Gaussian mixture models~\cite{bilmes1998}, latent linear models~\cite{jolliffe2014_llm}, sparse linear models~\cite{mairal2014}, ensemble learning~\cite{rokach2010}, directed graphical models~\cite{wainwright2008}, undirected graphical models~\cite{blake2011}, clustering~\cite{jain2010}, Gaussian processes~\cite{williams2006} , Dirichlet processes~\cite{teh2011}, and deep learning~\cite{bengio2013}.

The main contributions of this paper are: 
\begin {enumerate*} [label=\itshape\alph*\upshape)]
\item extensive survey of recently published hyperspectral analysis methods, \item categorization of each method by remote sensing task and machine learning algorithm (which is neatly summarized in Table~\ref{tab:summary}), and \item exploration of current trends and problems along with future directions. 
\end {enumerate*} 

This paper is organized as follows. Sections~\ref{dataanalysis_task_section} and~\ref{machine_learning_section} provide brief backgrounds on hyperspectral image analysis tasks and machine learning terminologies, respectively, to make the readers familiar of the topics needed to understand the surveyed methods. In Section~\ref{section_algo}, we survey the recent machine learning-based hyperspectral image analysis methods found in literature. Section~\ref{section_open_issues} discusses open challenges in the field of hyperspectral data analysis, and Section~\ref{section_conclusion} concludes the paper.

\section{Taxonomy of hyperspectral data analysis tasks}
\label{dataanalysis_task_section}


\begin{figure*}
\centering
\includegraphics[width=0.7\linewidth]{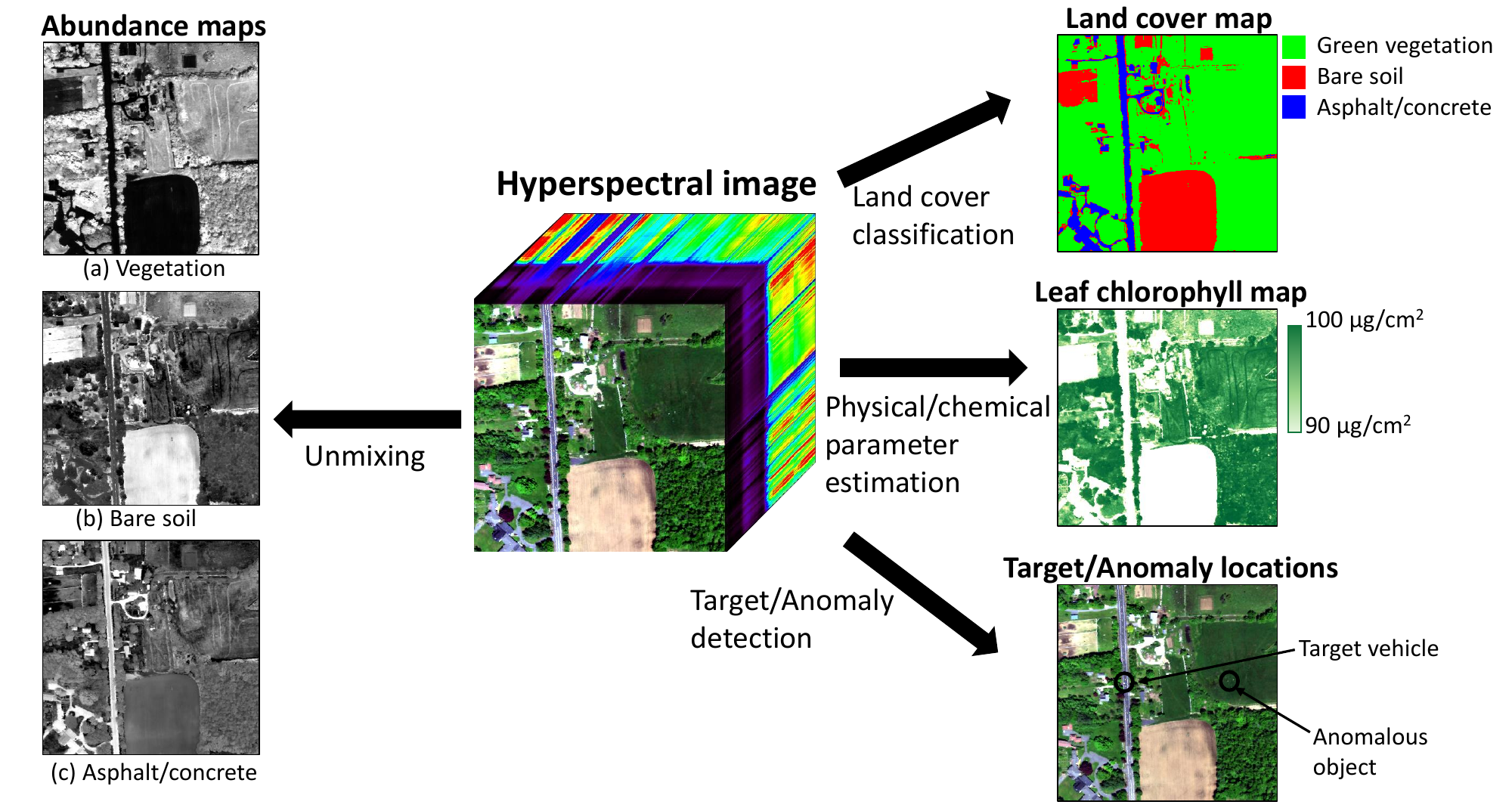} \\
\parbox{0.8\textwidth}{\noindent\rule{4.5cm}{0.4pt} \\
\scriptsize{The image was obtained from \cite{NEON_data}.}}
\caption{Hyperspectral image analysis tasks.}
\label{fig:hyperspectral_tasks_image}
\end{figure*}

The data analysis tasks for reflective hyperspectral images can be divided into four distinct groups: land cover classification, target/anomaly detection, spectral unmixing, and physical/chemical parameter estimation, as shown in Fig.~\ref{fig:hyperspectral_tasks_image}. These are further divided into sub-tasks. 
   
\subsection{Land cover classification}
\label{land_cover_classification_subsection}
Land cover classification~\cite{valls2014}, also called land cover mapping and land cover segmentation, is the process of identifying the material under each pixel of a hyperspectral image. The goal is to create a map showing how different materials are distributed over a geographical area imaged by the hyperspectral sensor. Common applications of land cover mapping are plant species classification~\cite{dalponte2013}, urban scene classification~\cite{dell2004}, mineral identification~\cite{murphy2012}, and change analysis~\cite{pu2008_change1}. The main advantage of using hyperspectral images to produce land cover maps is that hyperspectral images allow for the discrimination of land covers into finer class compared to other modalities, such as multispectral and panchromatic images, because hyperspectral images capture more information about the chemistry of the materials.  

Many land cover mapping methods require a prior knowledge about the types of materials present in the scene along with examples of the spectra belonging to different types of materials. This information is generally provided from the image pixels by an expert, collected using a spectrometer during field campaign of the study area, or adapted from a third party spectral library. However, there are also many land cover mapping techniques that require no prior information about the materials in the scene.

\subsection{Target/anomaly detection}
\label{target_detection_subsection}
Target detection~\cite{manolakis2002} is the task of finding and localizing target objects in a hyperspectral image, given a reference spectrum of the object. The target occurs very sparsely in the image, and can be composed of few pixels or even be smaller than a single pixel. Targets smaller than the size of a pixel are called sub-pixel targets. The reference spectrum is generally obtained from a spectral library. Generally, one or only few samples of reflectance spectrum of the target are only available. 

Anomaly detection~\cite{matteoli2010} is a related task with the objective of labeling anomalous objects in a hyperspectral image, without the prior knowledge of the object's spectrum. The size of the anomalous objects can also be in sub-pixel scale. Target and anomaly detection are widely used for reconnaissance and surveillance, and also in other areas like detection of special species in agriculture and rare mineral in geology~\cite{chang2013,dehnavi2017,makki2017}.

\subsection{Unmixing}
\label{unmixing_subsection}	
The energy captured by a pixel of a hyperspectral sensor is rarely reflected from a single surface of a single material. In airborne and space-borne imaging, the instantaneous field of view of a pixel (area covered by a pixel) on the ground is in meters, and it is highly likely that this area is covered by more than one material. For example, when imaging an agricultural land, the area under the pixel may contain vegetation and bare soil. Therefore, the measured spectrum is a combination of the spectra of different materials in the scene. This can be modeled as a linear mixture of spectra and the process is simply called linear mixing. Each material reflects energy in proportion to their coverage of the pixel area. Hence, the spectrum observed at the sensor is linear combination of spectra of the individual materials, weighted by their areal coverage. However, due to multiple scatterings of the light in the scene, the observed spectra is rarely a linear combination of spectra but, in fact, is a nonlinear combination~\cite{heylen2014}. Two common types of non-linear mixing models are bilinear mixing and intimate mixing. Bilinear mixing occurs when there are multiple reflections of the incident light on different materials on the scene. For instance, in forests the energy from the sun could get reflected from a leaf onto bare soil and then get reflected to the sensor. Intimate mixing occurs in fine mixtures, such as minerals, due to several multiple scattering from the particles in the mixture.

Hyperspectral unmixing is the process of recovering the proportions of pure material (called abundances) at each pixel of the image. The ``pure" material spectra are called endmembers. The endmembers present in a scene may be known a priori, or obtained from the image using an endmemeber extraction algorithm~\cite{chang2006,zortea2009}, or jointly estimated with the abundances. The  methods that require endmembers to be supplied are referred to as supervised unmixing and the methods that estimate endmembers simultaneously with the abundances are referred to as unsupervised in many literature. Applications of hyperspectral unmixing include mapping of green vegetation, non-photosynthetic vegetation and soil cover~\cite{meyer2015}, minerals exploration~\cite{rogge2014}, urbanization study~\cite{cavalli2008hyperspectral}, fire disaster severity study~\cite{robichaud2007}, and water quality mapping~\cite{olmanson2013}. 

\subsection{Physical/chemical parameter estimation}
\label{physical_parameter_estimation_subsection}	
Physical/chemical parameter estimation is the process of predicting contents of constituting chemicals or physical properties of materials, such as size, granularity and density of particles; structural characteristics and biomass of vegetation; and texture and roughness of surface, from reflectance spectra.  The chemical and physical properties of materials are manifested as spectral absorption features in the reflectance spectra, with the depth and the width of the absorption features being correlated to those parameters. Hence, it is possible to model physical and chemical properties of materials as functions of reflectance spectra. The physical and chemical parameters of vegetation are commonly referred as biophysical and biochemical parameters in literature. Some examples of applications of physical/chemical parameter estimation are prediction of leaf biochemistry~\cite{zhao2013_intro}, sand and snow grain size~\cite{painter2003,ghrefat2007}, vegetation biomass and structural parameter~\cite{cho2007,verrelst2012}, plant water stress~\cite{suarez2008}, and soil nutrient~\cite{anne2014}.

\section{Machine learning approaches}
\label{machine_learning_section}

Machine learning algorithms attempt to predict variables of interest by learning a model from data. This section provides a brief background on machine learning techniques and terminology that will be used to describe methods in remainder of the paper. As general references, the books by Kevin Murphy~\cite{murphy2012_book} and Christopher Bishop~\cite{bishop2006} provide a complete and detailed coverage of machine learning techniques.

\subsection{Types of learning}
Based on the type of learning, the machine learning methods can be broadly categorized into five groups as supervised learning, unsupervised learning, semi-supervised learning, active learning, and transfer learning.

In supervised learning~\cite{hastie2009supervised}, the relationship between the input and the output variables is established using a set of labeled examples, i.e., the examples for which the corresponding output variable values are known. The problem is called regression if the output variable is real and called classification if the output variable is discrete.

Unsupervised learning~\cite{hastie2009unsupervised} discovers the structure or the characteristics of the input data using unlabeled examples (examples for which corresponding output values are unavailable). For instance, k-means is an unsupervised learning algorithm that clusters the input data into homogeneous groups. The principal component analysis (PCA) is another unsupervised learning algorithm that can be used to find an uncorrelated linear low dimensional representation of the input data.
 
Semi-supervised learning~\cite{zhu2009} utilizes the unlabeled data along with the labeled data to build relationship between the input and the output variables. The unlabeled examples are used to learn the structure of the input variables, so that this information can be exploited to better learn the input-output relationship using the labeled data.
 
Active learning~\cite{settles2010} iteratively selects examples from the unlabeled data for manual labeling, and adds them to the labeled training set. The picked examples are the ones deemed most important for improving the input-output predictive performance. This cycle is repeated until the model exhibits a desired performance. In this way, the goal of active learning is to produce results similar to supervised or semi-supervised learning methods with fewer training examples. 

Transfer learning~\cite{pan2010} utilizes the information learned from one problem to solve another problem. The tasks (output variable), the domains (input variables) or both can be different between the source and the destination problems. Hence, compared to traditional learning, transfer learning allows the task, the domain or their distributions to be different during training and testing. Domain adaptation is a subset of transfer learning in which the source and the destination have different domains. Similarly, multitask learning is a type of transfer learning where multiple related tasks are simultaneously learned with the objective of improving the performances on both tasks.

\subsection{Non-probabilistic and probabilistic models}

\subsubsection{Non-probabilistic}
	Non-probabilistic models produce point estimation of the output and do not model the probability distribution of the output. These models have a decision or a regression function that estimates the value of the output based on the value of the input. These functions have some free parameters which are estimated by minimizing some cost function during training, with the goal of learning the input-output relationship. Certain penalties might be enforced on the possible values of the parameters via regularization to control the complexity of the model or to encourage certain properties in the solution.
\subsubsection{Probabilistic}
	The probabilistic models infer the probability distribution of the output. The generative probabilistic models consider both the input and the output as random variables, and model a joint distribution of the input and the output variables. In contrast, the discriminative probabilistic models consider the input to be deterministic and the output to be random, and model the distribution of the output variables as a function of the input variables, i.e., they model conditional distribution of the output given the input. In other words, the generative model learns the process by which the input and the output are generated, while the discriminative model only learns how to predict the output when the input is given. In probabilistic terms, generative model learns $p(x,y)=p(y|x)p(x)$ while discriminative model only learns $p(y|x)$, where $x$ and $y$ are input and output respectively. Generative models have the advantage of being able to generate samples of the data, however discriminative models typically perform better than generative models for classification and regression as it requires larger number of samples to learn a generative model (because they need to model $p(x)$ along with $p(y|x)$ to learn $p(x,y)$). The parameters in probabilistic models are also considered to be random variables, whose point values or distribution are to be inferred from the data. The parameters can be learned by maximum likelihood estimation, maximum a posteriori estimation, or Bayesian inference.  
	\begin{itemize}
	\item The maximum likelihood estimate (ML) estimation, makes point estimates for the parameters by maximizing the likelihood (probability) of observing the data given the parameters ($p(y|\theta)$ where $\theta$ represents parameters). The maximum a posteriori (MAP) estimation finds point estimates for the parameters by maximizing the probability of observing the parameters given the data (or posterior probability of parameters, $p(\theta|x)$). The ML estimation is equivalent to minimizing a cost function, and the MAP estimation equivalent to minimizing a cost function with regularization in non-probabilistic settings.
	\item The Bayesian inference finds the probability distribution of the parameters using the Bayes theorem, rather than just making point estimates. Exact Bayesian inference is intractable for many models, so different approximate inference techniques such as Laplace approximation, variational inference, and Markov chain Monte Carlo sampling have been developed. The main advantage of Bayesian inference over ML and MAP estimates is that Bayesian inference can properly model the prior believes about the model and handle uncertainties in the data and the model.  
	\end{itemize}

\subsection{Bias and variance}
The error in supervised learning models can be generally attributed to two sources: bias and variance. Bias is the error resulting from the model making wrong assumptions about the data. For instance, if a linear equation is used to model data whose input-output relationship is quadratic, the model will underfit and have high bias. It is characterized by high training error and high generalization (or testing) error. Underfiting occurs when the complexity of the model (the space of all the functions that the model can explain) is insufficient to represent the data. On the other side, variance is the error resulting from the sensitivity of the model to small changes in training data. It occurs when the model tries perfectly fit all the training data points, rather than generalizing the trend. High variance arises when a model overfits the data, such that it has low training error but high generalization error. Overfitting typically occurs in complex models with large number of parameters when the amount of training data is small.

\subsection{Parametric and non-parametric models}
Models with fixed number of parameters have fixed complexity and are called parametric models. These models will underfit if the data complexity is greater than the model's complexity. For instance, a linear regression model which has fixed number of parameters (equal to the number of input features) will have high bias if fitted on quadratic data. In contrast, a non-parametric model can increase the number of parameters (and hence its complexity) with the available training data.  An example of a non-parametric model is the nearest neighbor classifier. In nearest neighbor, the predicted class of a test sample is the class of the most similar training sample. In this model, the training data itself are the parameters. Hence, increasing the number of training samples increases the number of parameters and the  complexity of decision boundary that can be modeled.

\subsection{Model selection and performance evaluation}
Model selection is the process of choosing the best model for a task from a set of candidates. A performance on the training data is not a good metric for choosing best model as it does not capture generalization performance.  The goal of learning is to accurately make predictions on new data, not on the training data. Therefore, models are evaluated on a separate set of independent samples called the validation set. In model selection, the set of candidate models do not have to be entirely different algorithms, but could be same algorithm under different hyperparameter settings. Hyperparameters are the variables that control the properties of the model and are typically set before training. For instance, the number of layers in a neural network and the number of clusters in k-means algorithm are hyperparameters. Whenever enough data is not available for building separate training and validation sets, k-fold cross-validation technique can be applied. In this, the whole dataset is randomly divided into k disjoint subsets (folds), and one subset is used for validation while remaining k-1 are used for training. This process is repeated k-1 more times until each subset is chosen once for validation. Then, the results from all the folds is accumulated. In the extreme case, the value of k could be set as high as the number of training examples to get leave-on-out-cross-validation, which is useful when the number of training examples is very small.

If validation set is used to tune the hyperparameters of a model, the performance on validation set cannot be used as proxy for generalization performance of the method, because the hyperparameters were fitted to maximize performance on validation set. In this case, a third independent, separate set of samples, called testing set, should be used for performance evaluation to get unbiased estimate of the generalization performance.

\subsection{Feature extraction and feature learning}
Feature extraction is the process of generating an informative and meaningful representation of the data. Raw data is rarely fed as input to machine learning algorithms, but is preprocessed to be converted to a form that best highlights the required information in the data and is best suited for the learning algorithm. The dimension of the features can be larger or smaller than the dimension of the data. Feature extraction can be as simple as a linear transform or be a highly non-linear transformation. For example, short-time Fourier transform features of audio is better for speech recognition rather than raw audio signal. 

Feature learning is the process of learning the data transformation in feature extraction process from data itself. It is better than the traditional approach of manually designing feature transformation, because it can automatically discover transforms that generates good set of features for the tasks at hand using data statistics. Feature learning can be supervised or unsupervised based on whether it learns features from labeled or unlabeled data, respectively. Supervised neural networks and dictionary learning are the examples of supervised feature learning while principal component analysis, clustering, and autoencoders are examples of unsupervised feature learning.

\section{Machine learning for hyperspectral analysis}
\label{section_algo}
In this section, we survey recently published machine learning-based hyperspectral image analysis methods. Each subsection discusses methods that utilize a particular type of machine learning algorithm. The categorization of the machine learning algorithms is loosely based on the one used in Kevin Murphy's book~\cite{murphy2012_book}. Within each subsection, the methods are further grouped by the hyperspectral data analysis tasks. We survey 205 different methods in total. The number of methods surveyed for each type of machine learning algorithm and each image analysis task is listed in Table~\ref{tab:num_methods}.

\begin{table}
\small
\centering
\caption{The number of methods surveyed of each image analysis task and each machine learning algorithm.}
\label{tab:num_methods}
\begin{tabular}{lccccc}

\toprule
                            & \multicolumn{4}{c}{Image analysis tasks} & \\
\cmidrule(lr){2-5}
ML algorithms & Classification & Target & Unmixing & Parameter & All \\
\midrule
Gaussian models & 4 & 2 & 0 & 0 & 6 \\
Linear regression & 0 & 0 & 2 & 4 & 6 \\
Logistic regression & 9 & 0 & 0 & 0 & 9 \\
Support vector machines & 21 & 4 & 4 & 3 & 32 \\
Gaussian mixture models & 8 & 2 & 0 & 0 & 10 \\
Latent linear models & 9 & 0 & 1 & 5 & 15 \\
Sparse linear models & 9 & 1 & 8 & 0 & 18 \\
Ensemble learning & 15 & 1 & 0 & 4 & 20 \\
DGMs & 0 & 0 & 9 & 0 & 9 \\
UGMs & 19 & 0 & 5 & 0 & 24 \\
Clustering & 6 & 0 & 0 & 0 & 6 \\
Gaussian processes & 5 & 0 & 1 & 5 & 11 \\
Dirichlet processes & 1 & 0 & 4 & 0 & 5 \\
Deep learning & 33 & 1 & 0 & 0 & 34 \\
\midrule
All & 139 & 11 & 34 & 21 & 205 \\
\bottomrule
\end{tabular}
\scriptsize{
\setlength\tabcolsep{6pt}
\begin{tabular}{ll}
ML: Machine learning & \\
Target: Target detection & Parameter: Physical parameter estimation \\
DGM: Directed Graphical Model & UGM: Undirected Graphical Model
\end{tabular}
}
\end{table}

\subsection{Gaussian models}
Multivariate Gaussian models are the basis for most classical algorithms for land cover classification and target detection. A popular hyperspectral land cover classification algorithm is the quadratic discriminant analysis, also known as Gaussian maximum likelihood classifier or just maximum likelihood classifier~\cite{dalponte2013}. It is a discriminative model where the class conditional distribution of the data is assumed to be described by a multivariate Gaussian distribution, with the mean vectors and the covariance matrices estimated using maximum likelihood. A special case where all of the class covariance matrices are assumed to be the same is called linear discriminant analysis~\cite{bandos2009,li2011_kernel}. 

Gaussian models have also been extensively used in hyperspectral anomaly and target detection. The Mahalanobis distance detector~\cite{chang2002} models the pixel values of a hyperspectral image using a multivariate Gaussian, and labels the pixels having low likelihood under this distribution as anomalies. The Reed-Xiaoli (RX) detector~\cite{chang2002,matteoli2014} extends this by modeling only the neighborhood around the test pixel by a Gaussian distribution, not the entire image. Common target detection algorithms, such as spectral matched filter and adaptive cosine detector~\cite{manolakis2003,truslow2014}, also assume Gaussian distributions for the target and the background pixels.

Gaussian models can also be found as components in more advanced algorithms. For instance, the classification method by Persello et al.~\cite{persello2012}, which performs both active learning and transfer learning, utilizes Gaussian models. In this method, the class probabilities of the data were modeled by Gaussian distributions and query functions defined over the class probabilities were used to iteratively remove examples belonging to the source dataset from the training set and add examples belonging to the target dataset to the training set.

\subsection{Linear regression}
Linear regression is a widely used method for hyperspectral data analysis. It has been applied to physical parameter estimation and unmixing problems. Linear regression is a supervised method that learns a linear relationship between a set of real input variables and a output variable by modeling the output variable as the weighted sum of input variables plus a constant. In physical parameter estimation, it is used to relate the parameter of interest with the spectral reflectance values or features derived from the spectra~\cite{wang2008}. Some of the common features used are spectral derivatives~\cite{treitz1999}, tied spectra~\cite{kokaly2009}, and continuum removed spectra~\cite{kokaly1999}. Most of the studies use step-wise regression technique to select bands that have higher correlation to the parameter of interest. In step-wise regression, bands are one-by-one added or removed from the predictive model depending on whether their presence increases or decreases the predictive performance. When used for linear unmixing, the reflectance of observed spectra at each band is modeled as a weighted sum of reflectance of the endmembers at that band, with the weights being constant for all bands and corresponding to the abundances~\cite{heinz2001}. Using data transformations, some of the non-linear unmixing problems, such as bilinear mixtures and Hapke mixtures, can be solved using linear unmixing framework~\cite{heylen2015}.

\subsection{Logistic regression}
Logistic regression is a discriminative model primarily used for land cover classification in remote sensing. It models the class probability distribution as the logistic function of weighted sum of input features. It has been primarily used for pixel-wise classification, but as we will discuss later, logistic regression serves as a building component for more sophisticated algorithms that use ensemble learning, random fields, and deep learning. Logistic regression can perform classification with band selection using step-wise learning procedure~\cite{cheng2006} or using sparsity regularizer on the weights~\cite{zhong2008_logistic,pant2014,wu2015}. The sparsity regularizer forces many weights to be equal to zero during training, thus removing the corresponding bands from the model and keeping only the relevant bands. For improved performance, logistic regression have been trained on features derived from hyperspectral data. In \cite{khodadadzadeh2014}, the squared projections on subspaces derived from class-specific spectral correlation matrices were used with logistic regression. Qian et al.~\cite{qian2013} have proposed using 3D discrete wavelet transforms to obtain texture features from hyperspectral data cube for classification, and using mixture of subspace sparse logistic classifier to build a non-linear classifier. The 3D discrete wavelet transform based features have advantage of capturing both spatial and spectral contextual information of the scene. Spatial context can be also be incorporated to logistic regression by using morphological features~\cite{huang2014}.

Semi-supervised methods using logistic regression have also been proposed. These methods label the unlabeled data using heuristics and augment the training set with them. In \cite{dopido2014}, unlabeled pixels within the 4-neighborhood of labeled pixels were assigned the class of the labeled pixel and added to the training set, and in \cite{li2013}, the class labels of the unlabeled pixels were predicted by a Markov random fields based segmentation technique~\cite{li2012a} and added to the training set. 

\subsection{Support vector machines}
Support vector machines (SVMs) are the most used algorithms for hyperspectral data analysis~\cite{mountrakis2011support}. They have been successfully applied to all data analysis tasks (land cover classification, target detection, unmixing, and physical parameter estimation). SVM generates a decision boundary with the maximum margin of separation between the data samples belonging to different classes. The decision boundary can be linear, or be non-linear through the use of kernels~\cite{scholkopf2002}. Using kernels, the data can be projected into higher dimensional space where a linear decision hyper-plane is fitted, which in turn is equivalent to fitting a non-linear decision surface in the original feature space. The Gaussian radial basis function kernel is used by majority of the hyperspectral SVM algorithms, however several kernels specifically designed for modeling hyperspectral data~\cite{mercier2003support, fauvel2006evaluation, schneider2010gaussian, gewali2016novel} have been proposed. Since their introduction to the hyperspectral remote sensing in Refs.~\cite{huang2002,melgani2004}, SVM have been considered state-of-the-art classifier for land cover mapping.

The most accurate SVM-based land cover mapping methods utilize spatial-spectral features, such as extended morphological (EMP) features~\cite{benediktsson2005, fauvel2008}. The EMP features are generated by applying a series of morphological opening and closing operations with structural element of different sizes on principle component bands of the hyperspectral image. It has been shown in \cite{fauvel2008} that appending features generated by discriminant analysis to the morphological features can further increase the accuracy. Feature selection has also been incorporated to hyperspectral classification with SVM. For instance, genetic algorithms can be utilized to select the bands and optimize the kernel parameters~\cite{bazi2006}. Similarly, a step-wise feature selection can be performed on the SVM~\cite{pal2010}. Semi-supervised SVM that can utilize unlabeled data for training have also been developed~\cite{chi2007}. Relevance vector machine (RVM)~\cite{tipping2001}, a Bayesian probabilistic classification algorithm related to SVM, has also been applied for hyperspectral classification~\cite{braun2012,demir2007,mianji2011}. 

Multiple kernel learning tries to find a convex linear combination of an optimized set of kernel functions with optimized parameter that best describes the data. It has been shown that SVMs with multiple kernel can outperform SVM with single kernel for hyperspectral classification~\cite{tuia2010learning, gu2012representative, wang2016discriminative}. Using EMP features, multiple kernel learning framework can be used for spatial-spectral classification~\cite{gu2016nonlinear, liu2016class, li2015multiple}. Kernels defined over the spectra of neighboring regions (square blocks of pixels~\cite{camps2006composite} or superpixels obtained via segmentation~\cite{fang2015classification} around the test pixel) have also been combined with the kernel defined over the spectrum of the test pixel to perform spatial-spectral classification with SVM.  

A multiple kernel learning based transfer learning/domain transfer approach that simultaneously minimizes the maximum  mean  discrepancy  between the source and the destination datasets  along with  the  structural risk functional of the SVM was proposed for classification in \cite{sun2013}. This method was found to be better than regular SVMs and other SVM-based transfer learning schemes. Similarly, an active learning based domain adaptation method with reweighting and possible removal of samples from the source dataset was introduced in \cite{persello2013}. The pixels in the source dataset misclassified by the SVM in each iteration were removed, while the target dataset pixels with the most uncertain class assignments (based on the votes of binary SVMs trained in one-vs-all approach) were manually labeled and added to the training set.

In Refs.~\cite{wang2009,gu2013}, binary class SVM classification was used for unmixing by assuming that the pixels lying on or separated by the max margin hyperplanes to be the pure pixels, and the pixels occurring within the margin to be the mixed pixels. The abundances of impure pixels was then given by the ratio of the distance from the margin to the margin width. Using one-vs-all scheme this method was extended for scenes with more than two endmembers. Another approach for SVM-based unmixing is to generate an artificial mixed spectra dataset with known abundances, and learn a SVM model to classify the proportion of each endmember present in a test spectra at single percentage increments~\cite{mianji2011b}. The artificial dataset can be generated by calculating the randomly weighted sum of spectra belonging to a set of classes chosen at random from a list of endmembers. In another study~\cite{villa2011}, probabilistic SVM was used to generate per pixel probability of the pixel belonging to an endmember. The pixels with high probability of belonging to any one endmember were considered to be pure pixels, while the remaining pixels were considered mixed. The abundances in the mixed pixels were calculated using linear unmixing. The mixed pixels were further divided into subpixels, with the subpixels arbitrarily assigned to one of the endmember classes in numbers proportional to the class abundances. Simulated annealing was then used to arrange the subpixels in each mixed pixels to have spatial smoothness. This produced a sub-pixel mapping of the scene. 

Anomaly and target detection have been performed using a SVM related algorithm called support vector data description~\cite{tax2004}. This method generates a minimum enclosing hypersphere containing all the training data. Kernel trick can be used to find minimum enclosing hypersphere in a transformed domain. For anomaly detection, any pixel falling outside of the hypersphere enclosing all the pixels in the image are considered to be anomalies~\cite{banerjee2006,khazai2011,gurram2011}. While for target detection, an artificial training set of target pixels can be created by adding multinomial Gaussian noise to the target reference spectra, and any test pixel falling inside of the hypersphere enclosing the artificial dataset can be labeled as a target~\cite{sakla2011}. Nemmour et al.~\cite{nemmour2006} mapped change in land cover of an area by training a SVM classifier on the concatenation of spectra from images collected at multiple dates.

Previously, SVM regression was applied to predict biophysical parameters from multi-spectral imagery ~\cite{bruzzone2005,camps_valls2006,bazi2007}. While these method are also applicable to hyperspectral data, some newer methods have been developed specifically for hyperspectral data. In Refs.~\cite{camps_valls2009}, a semi-supervised method that uses kernel matrix deformed by labeled and unlabeled data was proposed. Active learning approaches for biophysical parameter estimation that select new samples based on distance from support vectors and the disagreement between the pool of SVM regressors trained on different subsets of training data have been proposed in \cite{pasolli2012}. The idea of learning related biophysical parameter simultaneously, exploiting the relationship between them using multitask SVMs was introduced in \cite{tuia2011}. The multitask SVMs were found to be more accurate than the individual SVMs in predicting biophysical parameters.

\subsection{Gaussian mixture models}
\label{GMM_subsection}
Gaussian mixture models~\cite{bilmes1998} represent the probability density of the data with a weighted summation of a finite number of Gaussian densities with different means and standard deviations. They are commonly used to model data that are non-Gaussian in nature or to group data into finite number of Gaussian clusters. Gaussian mixture model is a good choice to model class conditional probability in maximum likelihood classifier when the image spectra do not show Gaussian characteristics~\cite{dundar2002_gmm,li2012_gmm,li2014_gmm}. Same is the case with anomaly and target detection algorithms which traditionally utilized Gaussian distribution to model pixel and background probability density~\cite{tarabalka2009_gmm}. 

Gaussian mixture model have also been used to cluster hyperspectral data. \cite{tarabalka2009_gmmb} used Gaussian mixture model followed by connected component analysis to segment the hyperspectral image into homogeneous areas. A related method called independent component analysis mixture model, which models cluster density by non-Gaussian density have also been applied for unsupervised classification of hyperspectral data~\cite{shah2007_gmm}.

The popular clustering algorithm k-means~\cite{arthur2007} is a special case of Gaussian mixture model clustering~\cite{bishop2006}. K-means starts with initial guesses for cluster centers, assigns all the data points to a cluster based on the distance to the cluster centers, calculates the mean of data in each cluster, and updates each cluster center with the mean of that cluster. The process of grouping data, calculating the mean and updating the cluster centers is repeated until convergence. The biggest issue with k-means is that it requires the number of clusters in the data to be known a priori. ISODATA~\cite{ball1965} is a method based on k-means that does not require the number of clusters to be known a priori and works by merging and splitting the clusters in every k-means iteration on the basis of the distance between the clusters and the standard deviation of the data in each cluster. The k-means and the ISODATA are widely used for unsupervised classification of hyperspectral data~\cite{baldeck2013_gmm,narumalani2006_gmm}. Unsupervised classification maps produced by them have been fused with the results of pixel-wise supervised classification to perform spatial-spectral classification~\cite{yang2010_gmm,tarabalka2009_gmmb}.

K-means have also been used for anomaly detection and dimensionality reduction. \cite{duran2007_gmm} performed anomaly detection was by labeling pixels which were distant from the cluster centers found by k-means or ISODATA as anomalous pixel and \cite{su2011_gmm} used the cluster centers obtained from k-means as features for classification.

\subsection{Latent linear models}
Latent linear models find a latent representation of the data by performing a linear transform. The common latent linear models used in hyperspectral image analysis are the principal component analysis (PCA)~\cite{jolliffe2014_llm} and the independent component analysis (ICA)~\cite{hyvarinen2004_llm}. The PCA linearly projects the data onto an orthogonal set of axes such that the projections onto each axis are uncorrelated. The projection onto the first axis captures the largest portion of variance in the data, the projection on to the second axis captures the second largest portion of variance in the data and so on, such that the axes towards the end do not capture any variance in the data but only represent the noise.  On the other hand, the ICA linearly projects the data onto a non-orthogonal set of axes such that the projections onto each axis are statistically independent as possible. For PCA, the number of data samples has to be greater than or equal to the dimensionality of the data. Similarly, for ICA, the number axes onto which the data is projected has to be smaller or equal to the number of data samples. The PCA is primarily used for dimensionality reduction in hyperspectral images. The PCA is applied to the spectra in the image and only the projections which explain a significant proportion of the variance are kept~\cite{shaw2002_llm}. Reducing dimensionality makes models less likely to overfit and also removes noise. Hence, it is widely used as a preprocessing tool for hyperspectral analysis~\cite{xia2014,chen2015,monteiro2007_llm,farrell2005_llm}. The minimum noise fraction (MNF) transform~\cite{lee1990_llm}, which whitens the noise in the image before applying the PCA, is generally preferred over the PCA when reducing the dimensionality of highly noisy images. The PCA and the MNF can be used to perform non-linear dimensionality reduction using their kernalized variants~\cite{fauvel2009_llm,nielsen2011_llm}. The spatial-spectral features can be obtained by applying morphological operations after the PCA or the MNF~\cite{plaza2009_llm}.

The partial least squares (PLS)~\cite{vinzi2010_llm} regression is a widely used method for physical parameter estimation from hyperspectral data and is closely related to the PCA. The PLS fits a linear regression by projecting the input and the output onto separate linear subspaces where the covariance between their projections are maximized. Carrascal et al./~\cite{carrascal2009_llm} showed that PLS performs better than the combination of PCA and linear regression for hyperspectral data. It has been successfully applied to predict physical quantities, such as, soil organic carbon~\cite{gomez2008_llm}, biomass~\cite{cho2007_llm}, nitrogen~\cite{hansen2003_llm}, and water stress~\cite{bei2011_llm}. 

The ICA can also be used to reduce the dimensionality of hyperspectral data. \cite{wang2006_llm} observed that better classification results can be obtained if the dimensionality of hyperspectral image is reduced using the ICA compared to using the PCA or the MNF. Apart from dimensionality reduction, the ICA have also been used for unmixing~\cite{nascimento2005_llm} and unsupervised classification~\cite{du2006_llm}. These methods assume that ICA axes are the endmembers and the projections are the abundances. Mixture model using ICA have been also proposed for unsupervised classification~\cite{shah2007_gmm}. Similar to the PCA and the MNF, spatial-spectral features have been generated using morphological profiles on the image after ICA~\cite{dalla2011_llm}.

\subsection{Sparse linear models}
Linear sparse models~\cite{mairal2014} model the observed output to be the weighted linear combination of the elements (atoms) of a large dictionary with the restriction that most of the weights are equal to zero while the remaining few weights have significant magnitudes. The sparsity on the values of weights is imposed by using a a sparse prior in probabilistic setting and a sparse regularizer in non-probabilistic setting. The dictionary can be supplied manually or be learned from the data itself. When a dictionary is learned from the data, it is automatically able to capture the data statistics. The linear sparse model is widely used for unmixing, because its formulation resembles to the linear mixing model, with the abundances being the weights and the endmembers being the dictionary elements. Iordache et al.~\cite{iordache2011} have proposed the use of sparse linear model with a spectral library as dictionary, to linearly unmix images using $L_1$ regularizer on the weights. This method is able to automatically select a subset of spectra in the spectral library as endmembers for each pixel of the image. Note that the regular least squares cannot be used in this kind of settings, since the number of spectra in a spectral library is much larger than the number of bands in a hyperspectral image. In this problem, sparsity is imposing additional constraints on a ill-conditioned problem to make it solvable. A modified spatial-spectral version of this method~\cite{iordache2012}, additionally imposes the spatial contextual information by applying total variational regularization, i.e., minimizing the $L_1$ norm of the endmember-wise abundance differences between the neighboring pixels. 

A multitask spatial-spectral extension to the same method, where sparsity is imposed to all the pixels of an image simultaneously to force the neighboring pixels to be composed of same endmembers, using $L_{2,1}$ norm on the abundance matrix was proposed in Refs.~\cite{iordache2014,iordache2014b}. In all these methods, non-negativity constraint on the abundances was imposed during optimization, but the sum-to-one constraint was not applied. A hierarchical Bayesian approach to sparse unmixing was introduced in \cite{themelis2012}. Zero mean Laplace prior, estimated by a truncated Gaussian distribution, was used as prior on the abundance for sparsity and non-negativity constraints, and a deterministic heuristic to impose sum-to-one constraints was suggested. In \cite{castrodad2011}, multiple spectra belonging to each endmember classes were added one by one to the dictionary until there was no gain in reconstruction accuracy. The authors also proposed using non-local coherence regularizer that promotes coherence between coefficients corresponding to the atoms belonging to the same endmember class along with local neighborhood coherence regularizer.  

There is also a non-regularization based sparse endmember extraction and unmixing technique in literature. It first performs fully constrained least squares unmixing, and then iteratively removes endmembers with smallest abundances greedily until the required accuracy and sparsity were obtained~\cite{greer2012}. This method was shown to perform better that $L_1$ regularizer based sparse unmixing methods in the experiments. Manifold based regularizers have also been used for sparse unmixing with the assumption that hyperspectral data is sparse in a lower dimensional non-linear manifold in a high dimension space~\cite{lu2013}.

Classification can be performed with linear sparse models by either using the sparse representation as features for a classifier~\cite{charles2011,du2015} or by selecting the class that has minimum class-wise reconstruction error~\cite{haq2012,chen2011,chen2013,castrodad2011}. In \cite{charles2011}, dictionary learning  was used to infer sparse representation of the spectra, which was used as feature for SVM classification. In the experiments, it was found that using sparse feature performed better than using the raw spectra or the principal component analysis (PCA) transformed spectra. A compressed sensing based deblurring method to reconstruct hyperspectral signal from multispectral measurement was also introduced in this work. Similarly, in \cite{du2015} dictionary learning with total variation regularizer was used to learn discriminative spatial-spectral sparse representation jointly with the sparse multinomial logistic regression trained on them. 

The sparse reconstruction based classification algorithms learn the sparse representation of a test example using a dictionary containing examples of all classes, and then reconstruct that example using only dictionary atoms belonging to a specific class. The class with minimum reconstruction error is the predicted class for the test example. These methods use basis pursuit~\cite{castrodad2011}, greedy pursuit~\cite{chen2011} or homotopy based algorithms~\cite{haq2012} to obtain $L_1$ regularized sparse coefficients. Kernelized versions of sparse reconstruction algorithm can be used to construct dictionary in the higher dimensional kernel space~\cite{chen2013,liu2013}. In Refs.~\cite{chen2011,chen2013} spatial contextual information was utilized by enforcing that the Laplacian operation on the reconstructed image be equal to zero, and by optimizing joint sparsity that promotes neighboring pixels to be composed of the same dictionary elements. Similarly, a spatial-spectral classification method in which the surrounding pixels were weighted based on their similarity to the test pixel and reconstructed jointly with the test pixel using sparse model with spatial coherence regularizer was introduced in \cite{zhang2014}. 

A new set of sparse code can be learned from the learned sparse code. This process can be repeated multiple times, with a new set of sparse code being learned from the sparse code learned in previous step, to obtain deep sparse code. A superpixel guided deep spatial-spectral sparse code learning method was published in \cite{fan2017_sparse}. In this method, first the image was segmented into superpixels and sparse code was computed for spectrum at all the pixels of image. The learned sparse codes were averaged over the superpixels and assigned to all the pixels inside the superpixels. This process was repeated multiple times to generate a stack of features which were concatenated and classified using a support vector machine.   

Chen et al.~\cite{chen2011b} performed target detection by formulating the problem as a two class classification problem and using a previously proposed spatial-spectral sparse classification technique~\cite{chen2011}. Their method generated target training examples by running MODTRAN~\cite{berk1987} with randomly varying parameters on the reference target reflectance spectra, and used randomly selected pixels from the test image as training examples for the background. This method was shown to out-perform spectral match filters, matched subspace detectors, adaptive subspace detector, and SVM based binary classification.

\subsection{Ensemble learning}
Ensemble learning~\cite{rokach2010} is a supervised learning technique of merging the results from multiple base predictors to produce a more accurate result. The outputs of the base predictors must be diverse and uncorrelated for ensemble learning methods to produce superior results. It has been applied successfully for hyperspectral classification. There are basically three types of ensemble learning approaches: bagging, boosting, and random forest. Bagging (also called bootstrapping) combines the results from multiple predictors trained on randomly sampled subsets of training set, where as boosting combines the results from multiple separate weak predictors trained on the whole training set. Random forests is bagging of decision/regression trees with random selection of features (also called feature bagging). 

AdaBoost~\cite{freund1995}, an adaptive boosting technique, has been widely used to build robust hyperspectral classifiers. It was used with support vector machines (SVMs) trained on clusters of bands~\cite{ramzi2014}, multiple kernel support vector machines trained on screened training samples~\cite{gu2016}, support cluster machines with different number of clusters~\cite{chi2009}, and linear and quadratic decision stumps trained on randomly selected features~\cite{kawaguchi2007} . Random forest and other random feature subspace based ensemble learning methods are considered more attractive for hyperspectral data as they use a reduced feature set to learn each ensemble member, which makes them less prone to overfitting. A bias-variance analysis in \cite{merentitis2014} found that random forest with embedded feature selection and Markov random field (MRF) based post-processing are best suited for hyperspectral data. 

A random subspace SVM that trains multiple SVM classifiers on random subsets of bands and combines the result of each SVM on the test spectrum via majority voting was proposed in \cite{waske2010}. This method performed better than regular SVM and random forest, particularly when very few training examples were available. This method was further improved by optimizing every random subspace, by selecting the optimal bands using genetic algorithm with Jeffries-Matusita (JM) distance as fitness function in \cite{chen2014}. An adaptive boosting technique for random subspace SVM which jointly learns the kernel parameters of the SVMs and the coefficients of adaptive boosting was formulated in \cite{gurram2013}. Ham et al.~\cite{ham2005} introduced random forest algorithm to hyperspectral classification, and also proposed a novel idea of optimally selecting bands in each random subspaces projection using simulated annealing. Their method was extended to perform transfer learning by using the class hierarchy learned on the source image, when making predictions in the test (destination) image with few or no labeled data~\cite{rajan2006}. Transfer learning version of AdaBoost algorithm, TrAdaBoost~\cite{dai2007}, was used to reweight the samples from source image in the training set of a SVM classifier after unlabeled pixels in the target image were manually labeled and added to the training set in \cite{matasci2012}.

Extreme learning machines (ELM)~\cite{huang2011} are three layered feed-forward neural network, where the weights from input layer to hidden layer are assigned at random and the weights from the hidden to the output layer are learned. They were introduced to hyperspectral classification as base classifier for bagging and AdaBoost in \cite{samat2014}. They were found to be very fast compared to the methods using SVM as base classifier, while provide similar accuracy. The authors also proposed using extended morphological map (EMP) features for spatial-spectral classification with ELM.  Algorithms that dynamically select a separate subset of classifiers from an ensemble for each test pixel, considering the validation performance of pixels similar and near to the test pixel, have been proposed~\cite{damodaran2015}. This method used the SVM and the ELM as base classifiers, and were more accurate than similar methods with fixed set of ensemble members. 

A rotation forest~\cite{rodriguez2006} is a classifier which builds ensemble members by dividing the features of the training data into random groups, performing rotational transform on each group, and then collecting all the rotation vectors into a single transformation matrix which is used to rotate the entire training set before training a decision tree. The study by Xia et al.~\cite{xia2014}  applied rotation forest for hyperspectral data, with rotational transformations generated by principal component analysis (PCA), minimum noise fraction (MNF), independent component analysis (ICA), and Fisher's discriminative analysis and showed their method to outperform bagging, random forest, and AdaBoost. The authors later proposed improving the classification performance by exploiting spatial context using Markov random fields~\cite{xia2015b} and extended morphological map features~\cite{xia2015}.  

Peerbhay et al.~\cite{peerbhay2015} applied random forest classifier to identify anomaly by training a binary classifier on synthetic dataset which considered the image pixels to be the training samples of the non-anomaly class and the samples from empirical marginal distribution of image pixels to be the training samples of the anomaly class. Though primarily used for classification, random forest have also used for physical parameter estimation. In previous studies, random forest regression was used for band selection and retrieval of biophysical parameters, such as biomass~\cite{adam2014,mutanga2012}, nitrogen~\cite{rahman2013}, and water stress~\cite{ismail2010}.

\subsection{Directed graphical models}
Directed Graphical Models~\cite{wainwright2008}, also known as Bayesian networks, define a factorization of the joint probability of a set of variables over the structure of a directed graph. Each variable is represented by a node and the directed edges represent the conditional independence properties. Each variable in the joint distribution is considered to be conditionally independent given all its parents in the graph and the joint distribution is defined by the product of conditional distributions of all the variables given their parents. Bayesian networks have been primarily applied to hyperspectral unmixing. They have the advantages of providing uncertainty about the abundance estimates and probabilistically modeling the endmember spectral variability. Most methods follow a similar hierarchical Bayesian framework. They start with a linear mixing model or one of its nonlinear transforms, and assume  prior distributions over the abundances and the endmembers, and then use non-informative priors on the hyperparameters. The likelihood or the noise model used is mostly Gaussian. Since, exact Bayesian inference is not possible in these models, they use a Markov Chain Monte Carlo (MCMC)~\cite{andrieu2003}  method to estimate the posterior distribution of the abundances and sometimes also the endmembers. 

Based on the distributions used for priors and hyperpriors and the inference algorithm selected, different characteristics are obtained. A method to linearly unmix pixels when the endmembers of the scene are known was proposed in \cite{dobigeon2008}. The prior used on the abundances was a uniform sampling on a simplex to enforce non-negativity and sum-to-one constraints. This model was later extended to bilinear unmixing~\cite{halimi2011} and post-nonlinear unmixing~\cite{altmann2012}. In \cite{themelis2012}, the authors proposed the use of zero mean Laplace prior on abundance to promote non-negativity and sparsity on the retrieved abundance coefficients. For computational purposes, the Laplace prior was approximated by truncated Gaussian distribution. This model lacks sum-to-one constraint in its formulation, so the authors have suggested a heuristic to enforce this constraint. 

The variability in endmember spectra was addressed in \cite{eches2010}, where the endmember spectra were considered to be a Gaussian distributed vectors, with the means set to the endmember spectra extracted from the image and the covariance matrix learned from the data. In \cite{moussaoui2006}, joint estimation of the endmembers and the abundances distributions were made using a linear mixing model with independent Gamma priors on the mixing coefficient and the endmember reflectance values. The Gamma priors enforced positivity constraints on the mixing coefficient and the endmember reflectance values. Dobigeon et al.~\cite{Dobigeon2009} later added sum-to-one constraint to this model by replacing the independent Gamma prior on mixing coefficient by a Dirichlet prior. \cite{schmidt2010} proposed hardware and software implementation strategies to scale up these methods to large scale. Methods have also been devised to jointly estimate the abundances and the endmembers in bilinear and post-nonlinear methods~\cite{altmann2014}.

\subsection{Undirected graphical models}
\label{subsection_UGM}
The strong dependencies between neighboring pixels in hyperspectral images can been exploited for classification and unmixing using undirected graphical models (also called Markov random fields)~\cite{gewali2018tutorial}. Undirected graphical models (UGMs)~\cite{nowozin2011structured} are the probabilistic models that define a joint probability distribution of a set of random variables using the structure of an undirected graph, such that the joint distribution can be factorized over maximally connected sub-graphs, called cliques. Similar to directed graphical models, the nodes of the graph represent the variables while the undirected edges encode the conditional independence properties. In  UGMs, each node is conditionally independent to all of the the nodes given the neighbors of the node. The joint probability is equal to the normalized product of positive functions, called potential functions, defined over all of the cliques in the graph.

The most common type of undirected graphical model used in hyperspectral image analysis is grid-structured pairwise models. They have been widely used since their introduction to hyperspectral land cover mapping in \cite{jackson2002adaptive}. The graph structure used in these models is a grid, with pixel labels representing the nodes and an edge between every 4-connected or 8-connected neighbors. Two type of potential function is defined for these models, namely, the unary potential function and the pairwise potential function. The unary potential function is defined for each node and captures spectral information, while the pairwise potential function is defined for each edge and captures the spatial information by promoting neighboring similar pixels to be labeled the same class. It is common to use a pixel-wise classifier, such as Gaussian maximum likelihood classifier~\cite{jackson2002adaptive}, logistic regression~\cite{bores2011}, probabilistic SVM~\cite{tarabalka2010,xu2015}, Gaussian mixture model~\cite{li2014_gmm}, and ensemble method~\cite{merentitis2014}, to derive unary potentials. Ising/Potts based models~\cite{zhong2010,bores2011,tarabalka2010,xu2015,li2015} are primarily used for the pairwise potential. In these methods, two types of inference is generally performed--(a) maximum a posteriori (MAP) inference using algorithms such as iterated conditional mode~\cite{jackson2002adaptive}, simulated annealing~\cite{tarabalka2010}, and graph cuts~\cite{bores2011,xu2015,li2015} and (b) probabilistic inference (also called marginal inference) by using loopy belief propagation~\cite{zhong2010,li2013spectral}.

Tarabalka et al.~\cite{tarabalka2010} introduced a novel Potts-based spatial energy function that only created dependencies between the neighboring pixels if there was no intensity edge in between them. They found that SVM followed with edge-based Markov random fields performed better than SVM followed by non-edge based Markov random fields, which in turn was better than SVM followed by majority voting.  The conditional random fields (CRFs) are the Markov random fields that model conditional distribution by having their potential functions be parameterized by input features~\cite{sutton2011}. Studies~\cite{zhong2010,li2015} have successfully applied grid-structured pairwise CRFs to hyperspectral classification. CRFs are generally preferred over MRF because being discriminative model, they can better utilize training data for classification. However, to train a full CRF a large number of training examples are required because CRFs have much more parameters than MRFs. This is a problem for hyperspectral image analysis tasks, which are well-known for having limited ground truth. To tackle this problem, simpler formulations for the pairwise energy functions based on the similarity of the neighboring pixels have been proposed~\cite{zhang2012simplified,zhong2014support,zhao2015detail}. Band selection can also be jointly performed with the learning of a CRF to model land cover by applying Laplace prior on the CRF parameters as in \cite{zhong2008}. This helps in reduction of the complexity of CRF model and hence is better suited when there are limited training examples.  

All the methods discussed so far have only unary and pairwise potentials, and cannot express higher level relationship occurring between the different regions in the image. Higher order MRF/CRF have potential functions defined over a set with more than two nodes. Zhong and Wang proposed using robust $P^n$ model~\cite{kohli2009robust} for hyperspectral classification~\cite{zhong2011}. In this method, higher order potentials were defined over pixels inside each segment obtained by unsupervised segmentation of the hyperspectral image by mean-shift algorithm. This potential encouraged every pixel inside each segment to be assigned to the same class and was used along the unary and the pairwise potential functions. A heuristic approach is to use pairwise models to model segment labels directly and assign all the pixels in the segment to the same class. This approach is more rigid than $P^n$ model where all the pixels in the segment not necessarily have to be assigned to the same class. Similarly, \cite{golipour2015} integrated information from hierarchical segmentation map into pairwise energy function to incorporate higher order information into the pairwise MRF. 

A semi-supervised classification method that utilized MRF with semi-supervised graph priors on parameters was introduced in \cite{li2010}. Li et al.~\cite{li2011} proposed an active learning based classification method consisting of a pipeline of logistic regression followed by MRF, in which the unlabeled pixel having the largest difference in class probabilities for the top two most probable classes were selected for manual labeling and added to training set. Similarly, another heuristic approach for active learning that selected unlabeled pixels which were assigned to different classes by logistic regression  and the combination of logistic regression and MRF was proposed in \cite{sun2015_UGM}.

Eches et al.~\cite{eches2011} proposed using Markov random field and hierarchical Bayesian linear unmixing for simultaneous spatial-spectral unmixing and classification of a hyperspectral image. They modeled each pixel as the result of linear mixing of the endmembers with added zero mean Gaussian white noise. Potts-Markov model was used as prior for the class distribution over the image and separate sets of abundances prior were defined for each class. The abundances had softmax logistic regression as prior to enforce additivity and non-negativity constraints, and the coefficients of logistic regression had Gaussian prior with the mean and the variance distributed as normal and inverse gamma respectively. Bayesian inference on this model was performed using Metropolis-within-Gibbs algorithm. In the experiments, the proposed method performed better than pixel-wise Bayesian unmixing~\cite{dobigeon2008} and fully constrained least squares unmixing, but was more computationally intensive. The authors solved this problem by proposing an adaptive MRF which established relationships between homogeneous regions in image rather than all the pixels to reduce computational complexity in \cite{eches2013}. The new method used self-complementary area filter to segment the image into homogeneous regions as pre-processing. Potts-Markov model was used as prior to define class probability over these regions. Instead of using softmax logistic regression prior over class conditional abundances, this method used Dirichlet distribution with uniform prior on the parameters. This method performed as good as the previous method~\cite{eches2011} with lesser computational cost. 

A hierarchical Bayesian method for non-linear unmixing was proposed in \cite{altmann2014_UGM}. Each pixel in the image is modeled as sum of linear mixture of endmembers, a nonlinear term and Gaussian noise. All the pixels in the image were segmented into different classes by a MRF and the class conditional non-linear term was modeled by Gaussian processes. In the following paper~\cite{altmann2015}, the authors proposed to model the nonlinear term by a gamma MRF instead, to enforce the non-linear term to be non-negative and be not limited to finite levels. 

Sub-pixel mapping is the process of creating a classification map at scale smaller than the size of the pixel. The MRFs and CRFs have been used to create sub-pixel maps by first predicting the the content of coarse pixels, using methods like Gaussian likelihood estimated probability~\cite{wang2013} and linear unmixing~\cite{zhao2015}, then using this information to generate unary potentials for the sub-pixels inside all the coarse pixels, and finally using  MRF/CRF to globally assign labels to all sub-pixels in the image while promoting spatial smoothness.

\subsection{Clustering}
Clustering refers to grouping unlabeled data into homogeneous groups. Clustering algorithms have been primarily applied for unsupervised land cover classification of hyperspectral images~\cite{villa2013}, however they have been also used for band selection, semi-supervised classification, dimensionality reduction, and unmixing. The commonly used clustering algorithms are k-means~\cite{bishop2006}, ISODATA~\cite{ball1965}, meanshift~\cite{cheng1995}, affinity propagation~\cite{frey2007}, graph-based clustering~\cite{vonLuxburg2007}, and Dirichlet process mixture models~\cite{teh2011}. The k-means and the ISODATA methods are clustering methods that utilize mixture of Gaussians model and were discusses earlier in Section~\ref{GMM_subsection}.

Meanshift is an algorithm that can be used to find the locations of the modes in multi-modal distributions of data. It iteratively updates the estimates of the modes by the mean of all the data points weighted by a kernel function placed at the locations of the mode estimates. Meanshift does not require prior knowledge of the number of clusters unlike k-means, which is an advantage. \cite{huang2008} proposed using meanshift for unsupervised classification of a hyperspectral image. Then, they classified the mean spectra of each cluster using a supervised classifier and assigned all the pixels in the cluster with the predicted class.

Affinity propagation is a clustering technique that uses message passing between data points and does not require prior knowledge of the number of clusters as well. It has been used primarily for band selection by clustering the bands in an image and selecting a representative band for each cluster~\cite{jia2012,yang2013feature}. Graph based clustering techniques represent the structure of the data in graphs, with data points being nodes and the similarity between them being edges and express clustering as graph partitioning problem. Graph based clustering was used in \cite{camps2007} to perform semi-supervised land cover classification. Dirichlet process mixture models will be discussed in Section~\ref{DP_subsection}. Dirichlet process mixture models have been used for semi-supervised land cover classification~\cite{Jun2013}, unmixing~\cite{mittelman2012}, and endmember extraction~\cite{zare2010,zare2013}. 

\subsection{Gaussian processes}
Gaussian processes (GPs)~\cite{williams2006} are non-parametric models that assume all the observed and the unobserved data are jointly distributed by a multivariate normal distribution. The mean vector of the multivariate normal distribution is typically assumed to be a zero vector and the covariance matrix is estimated using a covariance function (same as kernel functions). Predictions are made using Bayesian inference. A popular covariance function for hyperspectral data analysis is the squared exponential function. The square exponential covariance function is called the Gaussian radial basis function in context of the support vector machines. GPs are primarily used for supervised learning, but can also be used for unsupervised learning. Most of the use of GP in hyperspectral data analysis has been in non-linear physical parameter estimation. The advantages of the GP over the traditional approach are that being probabilistic GPs provide a prediction confidence map; and being Bayesian they can handle uncertainties better and are less likely to overfit; and finally being non-parametric their complexity can grow with the amount of data to model highly non-linear functions.

GPs have been applied to predict leaf chlorophyll, fractional vegetation cover, and leaf area index in \cite{verrelst2012}. This study first showed that using GPs to predict biophysical parameters from common vegetation indices is more accurate than the standard technique of using linear regression. Then, they used GP to relate the raw spectrum to biophysical parameter and used step-wise backward elimination to select four bands whose reflectance could best predict the biophysical parameters. It was observed that GP with band selection performs the best. In the follow up study~\cite{verrelst2013} auto-relevance detecting (ARD) squared exponential covariance function was used to select bands to predict chlorophyll instead of stepwise regression.   

Both these techniques make assumption that the noise in each pixel is same, and it is independent of the signal. This assumption was relaxed in \cite{gredilla2014} by allowing the noise power to smoothly vary over observations using variational heteroscedastic Gaussian processes. This approach showed better performance than previous methods. In \cite{pasolli2012}, the authors proposed an GP-based active learning method for physical parameter estimation that utilized query functions based on the estimated covariance between labeled examples and unlabeled examples. The tutorial by Camps-Valls et al.~\cite{valls2016} covers biophysical parameter estimation using GPs from hyperspectral imagery in detail. In a different approach, Murphy et al.~\cite{murphy2014} used GP to accurately locate a feature in ferric iron spectra which occurs at around \SI{900}{\nano\metre} from only the shortwave infrared spectra (\SI{2000}{\nano\meter}-\SI{2500}{\nano\metre}), a commonly used spectral region in mineralogy. It was found that the estimated location of this feature highly correlated with the proportion of the mineral goethite.   

Gaussian processes can be employed for classification by using discrete likelihood functions. However, when using these functions, Bayesian inference does not have an analytical solution and approximate inference methods have to be used. In \cite{bazi2010}, Gaussian process classification was used to classify pixels in the image using squared exponential and neural network covariance functions, and logistic likelihood with Laplace and expectation propagation inference methods. In the experiments, their method outperformed the support vector machines. It was seen that the runtime of Gaussian process classifier became very high as the number of training sample was increased, and the authors proposed using approximate sparse GP methods, such as informative vector machines, for faster operation at the cost of some performance in applications where runtime is critical. An active learning based classification method that introduces three heuristics for Gaussian process classification of hyperspectral images was proposed in \cite{sun2015}.
 
A spatial-spectral classification scheme using GP was introduced in \cite{jun2011}. In this method, a set of GPs were used to model the per-class spatial variations in the reflectance in each band of the image pixels and for each class the mean reflectance estimate at each pixel was used as the mean of multivariate class-conditional normal probability distribution at that pixel. The observation angle dependent (OAD) function is a covariance function designed primarily for classifying minerals from hyperspectral data with GPs~\cite{melkumyan2009,schneider2014}. Five parsimonious Gaussian processes for modeling class-conditional distribution in quadratic discriminant classifier, when limited training samples are available, were introduced in \cite{fauvel2015}. Parsimony was enforced using constraints on eigen decomposition of the covariance matrices, assuming that the discriminating information lies on a lower dimensional subspace. 

Gaussian process latent variable model (GP-LVM)~\cite{lawrence2004} can be used for non-linear unmixing and endmember extraction~\cite{altmann2013}. This approach uses GP-LVM to perform non-linear dimensionality reduction to obtain abundances at each pixel of the image while enforcing positivity and sum-to-one constraints. Then, a GP regression model is trained to predict the pixel spectra from the estimated abundances. Finally, the endmember spectra are predicted by that GP regression model using pure abundances as input. This method works in reverse order compared to most other combined endmember and abundance estimation methods, in that, the abundance is estimated before extracting the endmembers.   

\subsection{Dirichlet processes}
\label{DP_subsection}
Dirichlet process (DP) ~\cite{teh2011}  is a Bayesian non-parametric model typically used to cluster data by modeling the data as a mixture of possibly infinite components. A set of infinitely many random variables are said to be distributed as Dirichlet process if the marginal of this joint is a Dirichlet distribution. When used for clustering, the number of clusters modeled by Dirichlet process can grow with the data, not requiring it to be set beforehand, unlike most clustering methods. Dirichlet process has been applied for classifying and unmixing hyperspectral data. 

In \cite{Jun2013}, a spatially adaptive, semi-supervised  DP based classification algorithm was proposed. This algorithm modeled the data by a DP based infinite mixture of Gaussian, with the Gaussian means given by spatially varying Gaussian processes. This method is capable of discovering new classes in the unlabeled samples in the training set, which is an uncommon feature in hyperspectral classification algorithms. A DP based joint endmember extraction and linear unmixing algorithm was proposed in \cite{mittelman2012}. This method used sticky hierarchical DP as spatial prior for the abundances in a Bayesian linear unmixing framework and Gibbs sampling to infer the posterior distributions of the endmembers and the abundances. 

Piecewise convex endmember (PCE) detection algorithm, proposed in \cite{zare2010}, models hyperspectral data as collection of disjoint convex regions. DP was used to determine the number and the location of these convex regions, and a maximum a posteriori estimate based method was used to estimate the Gaussian distributed endmembers in those regions. A stochastic expectation maximization algorithm was used to iteratively refine the extracted abundances and endmembers. A fully Bayesian version of this algorithm that uses Metropolis-within-Gibbs sampling for inference was introduced by the authors in \cite{zare2013}.

\subsection{Deep learning}
Deep learning~\cite{Goodfellow-et-al-2016,bengio2013} methods apply a hierarchy of non-linear transforms to the data with the goal of generating an abstract, useful representation. The growth in the development of graphical processing units (GPUs), availability of large datasets, and innovations in training deep networks such as dropout, rectified linear unit, residual learning, batch normalization, and dense connections have led to state-of-the-arts performances in the fields of computer vision, speech recognition, natural language processing, and many more engineering disciplines. Remote sensing researchers have also developed numerous deep learning based remote sensing data analysis methods which has produced  top performances. Currently, the main focus is on land cover classification task but in future we can expect deep learning to be used for other tasks as well. Due to its popularity, a tutorial~\cite{zhang2016} was also published recently to introduce deep learning to remote sensing researchers. In this section, we review the development of deep learning methods for hyperspectral analysis. 

Chen et al.~\cite{chen2014_deep} proposed classifying the features learned from stacked autoencoders by logistic regression for hyperspectral classification. Autoencoders~\cite{vincent2010} are the neural networks that map the input to a hidden layer of size smaller than that of input and then try to reconstruct the input signal at the output using hidden layer activations, so that the activations in the hidden layer provides a compact, non-linear representation of the input. The autoencoders can be stacked and fine tuned, such that the hidden representation of one is the input of the other, to learn deep features of the input. Chen st al.\ showed that the features generated by stacked autoencoder from hyperspectral image was better than that generated by commonly used feature extraction methods, such as principal component analysis (PCA), kernel PCA (kPCA), independent component analysis (ICA), non-negative matrix factorization and factor analysis. They extended this work by using deep belief network for learning deep representation in \cite{chen2015}. Deep belief networks are formed by stacking a kind of generative probabilistic neural networks, called restricted Boltzmann machines~\cite{le2008}. In both papers, three kinds of features were tested--spectral, spatial-dominated and spatial-spectral. The spectral features were generated by learning features using the raw spectra taken from the pixels of hyperspectral image as input. The spatial-dominated features used dominant principal components (PCA components) of the pixel neighborhood as the input to the feature extractor and the spatial-spectral approach used the concatenation of the raw pixel spectra and the dominant PCA components of the pixel neighborhood as the input to the feature extractor. They found that the performance of the spatial-spectral features was better than that of spatial-dominated features, which in turn was better than the performance of spectral features. Their method with spatial-dominated and spatial-spectral features also outperformed widely used spatial-spectral support vector machines (SVM). It was seen that the accuracy generally grew with the number of PCA components used up to a limit, and then remained constant. They found that accuracy is highly dependent on the depth of the network and suggest using schemes like cross-validation to learn the best depth. 

Liu et al.~\cite{liu2015} combined stacked denoising autoencoder (an autoencoder trained on noisy input) and super-pixel segmentation~\cite{li2012} for spatial-spectral classification. Their method used a three-layered stacked denoising autoencoder trained on pixels of an image to generate features to prepare a classification map. This classification map was then segmented using a super-pixel approach and all pixels in a segment was classified to a common class based on majority voting. Stacked autoencoders that jointly learn spatial-spectral features have also been proposed. Ma et al.\cite{ma2015} proposed the use of the PCA bands of the patch around the pixel and the pixel's spectra as joint input to stacked autoencoders for generating spatial-spectral features. They also showed that promoting pixels with smaller spectral angle to have same hidden layer representation in autoencoders trained on pixels and spatially averaging hidden layer representation of each autoencoder before stacking subsequent autoencoder can produce good spatial-spectral features in \cite{ma2016}.

A popular deep learning architecture for vision tasks is a convolutional neural network~\cite{krizhevsky2012} (CNN). Inspired by the mammalian visual system, these neural networks contain a series of convolution layers, non-linearity layers, and pooling layers for learning low-level to high-level features. In the convolution layers, rather than being fully connected to the input, each hidden layer unit is connected via shared weights to the local receptive field (pixel neighborhood for images) around the input. Non-linearity layer makes the activations non-linear function of the input. In the pooling layers, the responses at several input locations are summarized, via max operations, to build invariance to input translations.  Networks with one dimensional~\cite{hu2015}, two dimensional~\cite{romero2015}, and three dimensional~\cite{chen2016} convolutional layers have been developed for analyzing hyperspectral data. Methods with one dimensional convolution layer take spectra as input and learn features that capture only spectral information. \cite{hu2015} proposed a new five layered convolution neural network trained on the spectra for classification. This architecture contains an input layer, a convolution layer, a max pooling layer, a fully connected layer and an output layer. It was tested on two three hyperspectral datasets, and it was found that this method outperformed the SVM, and the existing shallow and deep neural networks, namely, two layer fully connected neural network, LeNet-5~\cite{lecun1995} and stacked autoencoder based method~\cite{chen2014_deep}. Li et al.~\cite{li2017_deep} devised a scheme to perform spatial-spectral classification using 1D CNN. They trained a 1D CNN on pairs of spectra to classify the common class if the spectra pair belong to same class and predict 0-th class otherwise. During testing, the test pixel and its neighboring pixels were paired one-by-one, fed to the network, and the class of the test pixel was found by majority voting. 

Generally, methods that use two dimensional convolution layers, first reduce the number of bands in the image by using a dimensionality reduction technique, such as PCA, and then apply two dimensional convolutional network on image patches to extract features. These methods reduce the number of bands to control the size of network, as larger network require more training data which is limited. In \cite{zhao2016}, the feature vector extracted from the last layer of a convolutional neural network, which was trained on the principal components bands of image patches, was concatenated with spectral feature vector obtained by applying a manifold based dimensionality reduction technique on spectra to obtain spatial-spectral features. Yue et al.~\cite{yue2015} generated spectral feature maps by performing outer products between different segments of the pixel's spectra and stacked it with spatial feature maps consisting of dominant PCA bands of image patch around the pixel to form spatial-spectral input, which was used to train a six layered convolutional neural network. In \cite{romero2015}, features generated by sparse unsupervised convolutional neural network, trained using Enforcing Population and Lifetime Sparsity (EPLS)~\cite{romero2015b} algorithm, showed better performance for classification than the features generated by the PCA and the kernel PCA with radial basis function kernel. Ghamisi et al.~\cite{ghamisi2016_deep} proposed a band selection method that chose the bands which maximized the accuracy of a 2D CNN over validation set using swarm particle optimization. Liang et al.~\cite{liang2016_deep} integrated 2D CNN and dictionary learning by learning sparse code from the last layers of a supervised 2D CNN, and using reconstruction loss to predict the class. They found that using dictionary learning was better than using support vector machines. 

The scale of the spatial features in a hyperspectral image is highly dependent on the instantaneous field of view and the geometry of the imaging system,  Zhao et al.~\cite{zhao2015_deep} proposed a multiscale convolutional autoencoder with three PCA bands to build scale invariance for classification. Their model passes the Laplacian pyramid of three dominant PCA bands of an image through two layers of convolutional autoencoder to generate spatial features which are concatenated with spectra and fed to logistic regression. Because of scale invariance, their method performed better than state-of-the-art spatial-spectral classification algorithms, the extended morphological profile method~\cite{benediktsson2005} and the multilevel logistic based method~\cite{li2010}. Aptoula et al.~\cite{aptoula2016_deep} proposed using attribute profile~\cite{mura2010_deep} as input to 2-D CNN for hyperspectral image as these profiles can capture structural information in an image at various scales.

Chen et al.~\cite{chen2016} proposed a three dimensional end-to-end convolutional neural networks to predict material class from the image patch around the test pixel. The term ``end-to-end" is used to denote networks that take raw input signal at the input layer and produce final output in the output layer and do not use any pre-processing or post-processing. Three dimensional convolutional networks are advantageous than two dimensional convolutional networks as they can directly learn spatial-spectral features as their filters span over both spatial and spectral axes. They also proposed using augmenting the training data with image patches generated by modeling the changes in illumination in the training examples and modeling linear mixing between the training examples to augment the training data. By augmenting the training data, they were able to train the proposed network without any dimensionality reduction of the image bands. Previous hyperspectral deep learning methods have relied on dimensionality reduction to decrease the number of bands in the image and hence limit the number of parameters in the model. The same idea of using 3D CNN for hyperspectral classification was again presented in \cite{li2017_deep_classification}. This work showed that 3D CNN outperformed baseline stacked autoencoder, deep belief network, and 2D CNN. Similarly, a 3D residual~\cite{he2016_deep} CNN was proposed in \cite{zhong2017_deep}. Residual networks have residual blocks which learn the difference between the target function and the input instead of the learning the whole target function which makes them more robust against overfitting when there are a large number of layers in the network.

Recently, generative adversarial networks (GANs)~\cite{goodfellow2014generative} were proposed for semi-supervised classification by Zhi et al.~\cite{zhi2017_deep}. GANs are generative models that learn to generate samples from the data distribution using two competing neural networks, one called the generator and the other called the discriminator. Zhi et al.\ trained a generator and a discriminator on spatial-spectral features obtained from hyperspectral image. Then, they added a softmax layer to the discriminator network at the end and fine tuned the network to perform classification. Their method was found to be effective when number of training examples were small.  

An architecture with two processing streams, a 1-D CNN stream for spectral information (trained on spectrum) and a 2-D CNN stream for spatial information (trained on image patch around the pixel), was proposed in \cite{zhang2017_deep}. It is different from methods that concatenate spectral and spatial features before classifier in that this method jointly optimizes the spectral and the spatial feature extractors. Similarly, Yang et al.~\cite{Yang2017_deep} performed transfer learning using a two stream spatial-spectral network. They trained the network using two fully annotated source images, fixed the weights of earlier layers, and fine-tuned final layers on test images. They found that transferring weights produces better results than training the entire network from scratch on the test image. Santara et al.~\cite{santara2017_deep} proposed an end-to-end network that selects groups of non-overlapping bands, processes them in identical parallel processing streams, and combines them to produce the final output.    

A fully convolutional network is an end-to-end deep network with only convolutional layers and no fully connected layers. It is trained to map an arbitrary sized image directly to its classification map. The method by Jiao et al.~\cite{jiao2017_deep} used FCN-8~\cite{shelhamer2017_deep} network pretrained on RGB ground based images to generate multi-scale features, which were then weighted and concatenated with spectrum to form spatial-spectral features for classification. Since, FCN-8 is trained on 3 bands (Red-Green-Blue), this method only use 3 dominant PCA bands of hyperspectral as input. In \cite{lee2017_deep}, a novel fully convolutional architecture was proposed for hyperspectral images.  The network was trained on hyperspectral image patches as input and corresponding ground truth map of the image patches as output. This network used a multi-scale convolutional filter bank to extract features at multiple scales in the first layer. The network also contained skip connections for residual learning~\cite{he2016_deep} and data augmentation (mirroring along horizontal, vertical, diagonal axes) to make optimization easier. Due to fully convolutional nature of the network, at test time the entire test image was passed into the network to generate classification map the whole image simultaneously.   
	
Tao et al.~\cite{tao2015} proposed using two layered stacked sparse autoencoders (an autoencoder which promotes sparsity of the activation in hidden layer during training) to learn multiscale spatial-spectral features in an unsupervised manner. Their method applied PCA on an hyperspectral image and extracted random square patches of different sizes to learn a set of autoencoders from the dominant principal component bands at different scales. Results from all the autoencoders were concatenated to obtained a final feature vector which was used for classification with a linear SVM. This was the first study that investigated the transferability of features in hyperspectral images. They found that features learned on separate source image can be as good as features learned on the test image itself. Similarly, Kemker and Kanan~\cite{kemker2017_deep} proposed using multi-scale ICA and stacked convolutional autoencoder to learn unsupervised spatial-spectral representation from images patches. Compared to \cite{tao2015}, they used a larger training set from multiple sensors for feature learning. On the other hand, studies~\cite{Yang2017_deep, mei2017_deep} have investigated the idea of pretraining a CNN on an annotated source hyperspectral image and then finetuning only the end layers on the test hyperspectral image to learn spatial-spectral representation in a supervised manner. A domain adaptation based method that learns features that are invariant to the difference in the distribution of the source image and the destination image was proposed in \cite{elshamli2017_deep}. Their method uses an end-to-end domain-adversarial neural network~\cite{ganin2016_deep} to learn features that maximizes the separability of material classes but minimizes the separability based on whether the features came from the source image or the destination image. It was found that this feature was better than feature learned from applying denoising autoencoder, PCA, and KPCA on source image and baseline deep networks trained on the source and the destination image.  
 
Even though all the algorithms discusses in this section have been classification algorithms, deep learning can also be used for other tasks. A deep and transfer learning based anomaly detector has been recently published in \cite{li2017_deep_anomaly}. In this method, a two-class one dimensional CNN was trained to detect the dissimilarity between two spectra. Training set was generated by selecting pairs of spectra from a fully annotated separate training image. The pair of spectra were assigned to 0-class if they belonged to same material and 1-class if they belonged to different materials. During testing, each pixel in the test image was compared to all its neighbors and the score obtained from the network was averaged. Anomalies were then detected by thresholding this score.   

Recurrent neural networks (RNN) are popular architectures for modeling sequential data. They contain feedback loops in their computation allowing the current output to be dependent on the current input and the previous inputs. This is different from all the networks we have discussed previously which used feedforward computations to produce output. Mou et al.~\cite{mou2017_deep} proposed using RNN to model pixel spectra in a hyperspectral image as a 1-D sequences for classification. They experimented with architectures based on two kinds of recurrent units, namely, long short-term memory~\cite{graves2005_deep} (LSTM) and gated recurrent unit~\cite{cho2014_deep} (GRN). They found that the GRN worked better than the LSTM for modeling hyperspectral data and both of the recurrent networks outperformed traditional approaches and baseline CNN. Similarly, Wu et al.~\cite{wu2017_deep} showed that a convolutional RNN~\cite{zuo2015_deep} (a network that has few convolutional layers followed by RNN) is better choice for spectra classification than LSTM and baseline CNN.

\section{Open Issues and Future Challenges}
\label{section_open_issues}

\subsection{Curse of dimensionality}
The high dimensionality of hyperspectral data is a well-studied problem in remote sensing. Some of the technique proposed to tackle it are dimensionality reduction~\cite{shaw2002_llm}, fusion of spatial information~\cite{fauvel2013}, transfer and multitask learning~\cite{Yang2017_deep,tuia2011}, and supplementation of training data with synthetic examples~\cite{chen2016}. However, there is still much more to explore about the nature of hyperspectral data. Two of the open questions about the characteristics of hyperspectral data are intrinsic dimensionality~\cite{lunga2014manifold} and virtual dimensionality~\cite{chang2018review}. Since, hyperspectral bands typically oversample reflectance spectra in many wavelengths, it is imperative that hyperspectral data lies in a smaller subspace within a higher dimensional space. The minimum number of dimensions required to carry all of the relevant information in an image is called intrinsic dimensionality. There is a need for methods that can extract a lower dimensional representation of hyperspectral data that can be applied on diverse images for a wide range of applications. Proper quantification and understanding of the intrinsic dimensionality would not only lead to development more efficient analysis algorithms, but also aid in the development of cost-effective, efficient sensors with fewer optimized bands. Virtual dimensionality is a related concept and is defined as the total number of possible unique spectral signatures. As discussed earlier, reflectance spectrum of a material is a linear or non-linear combination of its constituent "pure" materials' reflectance spectra. This means every reflectance spectrum can be categorized as pure (endmember) or impure (mixed). So, if there is a fixed, quantifiable intrinsic dimensionality of hyperspectral data, there should also be only a fixed number of pure materials. Virtual dimensionality is a relatively unexplored idea compared to intrinsic dimensionality and can have profound impact on hyperspectral data analysis if it is properly understood.  

\subsection{Robustness and reliability of models}
The grand challenge of hyperspectral data analysis is to build models that are invariant to the differences in factors such as, the time and season of image acquisition, site, platform, spatial resolution, spectral resolution, band sampling intervals, and sensor technology. It is an ambitious task as there are massive variability in the spectra when any one those factors is even slightly altered. Therefore, studies so far have mostly concentrated on developing models that work well for a particular image. These models are typically trained on the labeled pixels in the test image itself or on the ground spectra and ground truth information collected over the imaged site, and evaluated on the remaining unlabeled pixels. Nonetheless, the eventual goal is to build universal models which are operable under different conditions. 

Interest towards such models is beginning to grow in the remote sensing community, as seen by the contest problem in the 2017 IEEE Geoscience and Remote Sensing (GRSS) data fusion competition\footnote{\url{http://www.grss-ieee.org/community/technical-committees/data-fusion/}}. The task in the competition is to map the land use of new test cities using ground truth land use information of separate cites in the training images using multispectral data collected by different satellites at different seasons. Once the satellite based hyperspectral imaging becomes more mature, such datasets will be available for hyperspectral images as well and researchers will have opportunities to invent new methods for multi-city hyperspectral classification.

\subsection{Big data without ground truth}
After the initial cost of the instruments, it is often cheaper to obtain large quantity of hyperspectral images than to collect ground truth information for even a small area. For some of the ground truth information, such as chemical composition and physical properties of the materials that is needed for unmixing and physical/chemical parameter estimation, it is only practical to obtained ground truth for very few samples. This has lead to availability of large databases of images without ground truth or with very limited ground truth. With the increasing quality and miniaturization of sensors and their decreasing costs, this amount is only expected to grow at an exponential rate as unmanned aerial vehicle-based and satellite-based imaging becomes more popular. The availability of this massive amount of ground truth-less images raises an interesting question that whether they can be used to improve the existing methods and create new methods. Unsupervised deep learning could be a key to answering this question.

While the development of robust supervised deep learning methods is contingent on the development of large-scale datasets, unsupervised deep learning can learn robust representation of hyperspectral data using vast amount of already available unlabeled images. Studies have already investigated the transferability of learned unsupervised features between images of diverse scenes~\cite{tao2015,kemker2017_deep}. Deep generative models, such as generative adversarial networks (GANs) and variational autoencoders (VAs), look very promising for modeling unlabeled hyperspectral data. GANs and VAs could characterize the spectral variability by modeling the generative distribution of the spectra. Such generative models, in turn, can be used as prior for classification algorithms to make them invariant to spectral variabilities. If such generative models are conditioned on the set of physical/chemical parameters of the material, they could be used as data-driven forward models in lieu of physics-based radiative transfer models. These models could find usage in non-linear unmixing and physical/chemical parameter estimation. Similarly, the generative models conditioned on material class can also serve as a spectral library. GANs and VAs could also model the spatial prior of the land covers (similar to how GANs were used in \cite{luc2016semantic}). Deep learning based spatial prior could turn out to be better than Markov random fields based prior for land cover classification. GANs and VAs for hyperspectral images should also prove to be good for image processing tasks, such as pansharpening, superresolution, denoising, and inpainting.
 
\subsection{Lack of standardized datasets and experiments}
There is a lack of benchmarking datasets and experiments for hyperspectral analysis. Without a set standard procedure to evaluate methods under real-world scenarios, researchers and practitioners cannot make an educated choice in picking a right method for their problem. Since, different researchers have been using different datasets with different experimental conditions, it is virtually impossible to compare two methods proposed in two different papers. On top of that, many times it is difficult to reproduce a study because the implementation of the methods is not readily available. These factors stifle the possibilities of follow up work and adoption of the methods. Some of the current efforts to provide a platform for benchmarking different methods are the IEEE Geoscience and Remote Sensing society's (IEEE GRSS) Data and Algorithm Standard Evaluation website~\cite{dell2017ieee}, IEEE GRSS annual data fusion contest~\cite{debes2014_df,liao2015_df,moser2015_df,tuia2016_df}, and Rochester Institute of Technology's target detection blind test website~\cite{snyder2008development}.

Of the few available public hyperspectral datasets, most are for land cover classification. This has contributed to more methods being developed for land cover classification, than any of the other data analysis tasks. Researchers working on target/anomaly detection, unmixing, and physical/chemical parameter estimation generally use simulated data~\cite{duran2007_gmm,iordache2011,pasolli2012} or private datasets~\cite{kokaly2009,banerjee2006,ismail2010} to validate their methods, as there are few public datasets, e.g. Refs.~\cite{leafchem1999, pieters2000spectral,snyder2008development}, for these tasks. The current land cover classification datasets contain images that are small compared to real-life images and lack variability and diversity. Most of them were captured by same sensor as well. This raises a question that whether the current methods, which work great on these few and small images, will generalize to performing good on large real-world images. Due to all these reasons, building new public datasets should be one of the top priorities of the hyperspectral remote sensing community.

\begin{sidewaystable*}[ph!]
\fontsize{3}{3}\selectfont
\caption{Summary of all reviewed methods.}
\label{tab:summary}
\centering
\renewcommand{\arraystretch}{3}
\begin{tabularx}{\paperwidth}{@{}lXXXXXXX@{}}
\toprule
Methods & \multicolumn{2}{c}{Classification} & \multicolumn{2}{c}{Detection} & \multicolumn{2}{c}{Unmixing} & Parameter estimation\tabularnewline
\cmidrule(lr){2-3}
\cmidrule(lr){4-5}
\cmidrule(lr){6-7}
 & \multicolumn{1}{c}{Pixel-wise} & \multicolumn{1}{c}{Spatial-spectral} & \multicolumn{1}{c}{Target} & \multicolumn{1}{c}{Anomaly} & \multicolumn{1}{c}{Linear} & \multicolumn{1}{c}{Non-linear} & \tabularnewline
\midrule
Gaussian Models & \cite{dalponte2013,bandos2009,li2011}; transfer learning (\cite{persello2012}) &  & \cite{manolakis2003} & \cite{chang2002} &  &  & \tabularnewline
Linear Regression &  &  &  &  & \cite{heinz2001} & \cite{heylen2015} & band selection (\cite{wang2008,treitz1999,kokaly2009,kokaly1999})\tabularnewline
Logistic Regression & \cite{khodadadzadeh2014}; band selection (\cite{cheng2006,zhong2008_logistic,pant2014,wu2015}) & \cite{qian2013,huang2014}; semi-supervised (\cite{li2013,dopido2014}) &  &  &  &  & \tabularnewline
Support Vector Machines & \cite{huang2002,melgani2004,braun2012,demir2007,mianji2011,tuia2010learning, gu2012representative, wang2016discriminative}; band selection (\cite{bazi2006,pal2010}); semi-supervised (\cite{chi2007}); transfer learning (\cite{sun2013,persello2013}); change detection (\cite{nemmour2006}) & \cite{benediktsson2005, fauvel2008, gu2016nonlinear, liu2016class, li2015multiple, camps2006composite, fang2015classification} & \cite{sakla2011} & \cite{banerjee2006,khazai2011,gurram2011} & \cite{mianji2011b}; endmember extraction and sub-pixel mapping (\cite{villa2011}) & \cite{wang2009,gu2013} & semi-supervised (\cite{camps_valls2009}); active learning (\cite{pasolli2012}); multi-output (\cite{tuia2011})\tabularnewline
Gaussain Mixture Models & \cite{dundar2002_gmm,li2012_gmm,li2014_gmm}; unsupervised (\cite{tarabalka2009_gmmb,shah2007_gmm}); band selection (\cite{su2011_gmm}) & \cite{yang2010_gmm,tarabalka2009_gmmb} &  & \cite{tarabalka2009_gmm,duran2007_gmm} &  &  & \tabularnewline
Latent Linear Models & dimensionality reduction (\cite{shaw2002_llm,lee1990_llm,fauvel2009_llm,nielsen2011_llm,wang2006_llm}); unsupervised (\cite{du2006_llm,shah2007_gmm}) & feature extraction (\cite{plaza2009_llm,dalla2011_llm}) &  &  & \cite{nascimento2005_llm} &  & \cite{carrascal2009_llm,gomez2008_llm,cho2007_llm,hansen2003_llm,bei2011_llm}\tabularnewline
Sparse Linear Models & \cite{haq2012,castrodad2011}; feature extraction (\cite{charles2011}) & \cite{liu2013,chen2011,chen2013,zhang2014}; feature extraction (\cite{du2015,fan2017_sparse}) & spatial-spectral (\cite{chen2011b}) &  & \cite{iordache2011,themelis2012,greer2012,lu2013}; spatial-spectral (\cite{iordache2012,iordache2014,iordache2014b,castrodad2011}) &  & \tabularnewline
Ensemble Learning & \cite{waske2010,chen2014,gurram2013,ramzi2014,chi2009,kawaguchi2007,ham2005,damodaran2015,xia2014}; transfer learning (\cite{rajan2006}); transfer and active learning (\cite{matasci2012}) & \cite{merentitis2014,samat2014,xia2015b,xia2015} &  & \cite{peerbhay2015} &  &  & band selection (\cite{adam2014,mutanga2012,rahman2013,ismail2010})\tabularnewline
Directed Graphical Models &  &  &  &  & \cite{dobigeon2008,themelis2012,eches2010,moussaoui2006,Dobigeon2009,schmidt2010} & \cite{halimi2011,altmann2012,altmann2014} & \tabularnewline
Undirected Graphical Models &  & \cite{zhong2010,bores2011,tarabalka2010,xu2015,merentitis2014,li2015,zhong2011,golipour2015,li2014_gmm,li2013spectral,jackson2002adaptive,zhang2012simplified,zhong2014support,zhao2015detail}; band selection (\cite{zhong2008}); semi-supervised (\cite{li2010}); active learning (\cite{li2011,sun2015_UGM}); sub-pixel mapping (\cite{wang2013})  &  &  & spatial-spectral (\cite{eches2011,eches2013}); sub-pixel mapping (\cite{zhao2015}) & spatial-spectral (\cite{altmann2014_UGM,altmann2015}) & \tabularnewline
Clustering &  unsupervised (\cite{villa2013,huang2008,Jun2013}); band selection (\cite{jia2012,yang2013feature}); semi-supervised (\cite{camps2007}) &  &  &  &  &  & \tabularnewline
Gaussian Processes & \cite{bazi2010,melkumyan2009,fauvel2015}; active learning (\cite{sun2015}) & \cite{jun2011} &  &  &  & endmember extraction (\cite{altmann2013}) & \cite{gredilla2014,murphy2014}; band selection (\cite{verrelst2012,verrelst2013}); active learning (\cite{pasolli2012})\tabularnewline
Dirichlet Processes &  & \cite{Jun2013} &  &  & endmember extraction (\cite{Jun2013,mittelman2012,zare2010,zare2013}) &  & \tabularnewline
Deep Learning & supervised feature learning (\cite{mou2017_deep,wu2017_deep}); unsupervised feature learning (\cite{vincent2010,chen2014_deep,chen2015}) & unsupervised feature learning (\cite{chen2014_deep,chen2015,liu2015,zhao2015_deep}); supervised feature learning (\cite{hu2015,romero2015,chen2016,zhao2016,yue2015,ma2015,ma2016,li2017_deep_classification,aptoula2016_deep,zhang2017_deep,jiao2017_deep, lee2017_deep, zhong2017_deep, santara2017_deep, li2017_deep, liang2016_deep}); semi-supervised (\cite{zhi2017_deep}); supervised transfer learning (\cite{Yang2017_deep,mei2017_deep,elshamli2017_deep}); unsupervised transfer learning (\cite{tao2015,kemker2017_deep,elshamli2017_deep}); supervised band selection (\cite{ghamisi2016_deep})  &  & \cite{li2017_deep_anomaly} &  &  & \tabularnewline
\bottomrule
\end{tabularx}
\end{sidewaystable*}

\section{Conclusion}
\label{section_conclusion}
Over the years, machine learning has become the primary tool for analysis of hyperspectral images. With the literature booming with new methods, choosing the right method for a problem can become a daunting task. This paper addressed this issue by creating a catalog of published methods. We have summarized all of the discussed methods in Table~\ref{tab:summary}.

The current hot topic in hyperspectral data analysis is deep learning. Researchers have demonstrated deep learning to be the state-of-the-art for land cover classification~\cite{zhang2016}. Hyperspectral datasets are infamous for having small amount of ground truth data, while the deep learning methods are infamous for requiring large amount of ground truth data. Current methods get around this problem by training the network on labeled pixels or small patches of an image and testing on the remaining unlabeled pixels in the same image, instead of training and testing on separate images. They additionally employ regularization schemes, such as data augmentation and early stopping, to prevent overfitting. Since, the current models are learned on training sets consisting of spectra from a single image, which is typically small, these models cannot capture the spectral variabilities that occur due to differences in factors, such as illumination, atmospheric conditions, sun angle, viewing geometry, and resolution. If there were large-scale datasets with labeled images covering diverse scenes for training, it is highly likely that the networks would be able to learn a spectral variabilities resistant representation of the data, and  would perform well when tested on unseen new images. This highlights the fact that there is an urgent need of large-scale datasets for land cover classification. Such datasets could have as powerful of an impact to the field of hyperspectral image analysis as the development of ImageNet~\cite{deng2009imagenet} had for the growth of computer vision and deep learning.

Compared to land cover classification, other tasks have seen less attention from researchers developing machine learning-based hyperspectral image analysis algorithms, as demonstrated in Table~\ref{tab:num_methods}. This is primarily because it is more difficult to obtain sufficient ground truth information for these tasks. Target/anomaly detection, unmixing, and physical/chemical parameter estimation are as important, if not more important, than land cover classification. In fact, some tasks such as sub-pixel target/anomaly detection and unmixing are typically only possible using hyperspectral data, not by using other types optical images. Therefore, rather than mostly focusing on land cover classification,  hyperspectral algorithm researchers should break the current trend and start developing methods for other hyperspectral analysis tasks, and even try to propose and solve new tasks. As discussed in Section~\ref{section_open_issues}, generative deep generative models, such as generative adversarial networks and variational autoencoders, could have great impact on these tasks in the future. As of now, Bayesian methods, in particular hierarchical Bayesian unmixing models~\cite{eches2010,altmann2014_UGM} and Gaussian processes based models~\cite{verrelst2013,valls2016}, seem to be best suited for the task of unmixing and physical/chemical parameter estimation due to their flexibility, ability to handle uncertaininty in data, and capacity to perform good with limited training data. As for target/anomaly detection, classical statistical methods~\cite{truslow2014} are still the best choice since machine learning-based methods for these tasks are not well developed and tested. 

\bibliographystyle{plain}
\bibliography{bibtex/bib/tutorial,bibtex/bib/clustering,bibtex/bib/LLM,bibtex/bib/GMM,bibtex/bib/UGM,bibtex/bib/deep_learning,bibtex/bib/introduction,bibtex/bib/linear_gaussian,bibtex/bib/SVM,bibtex/bib/DGM,bibtex/bib/logistic,bibtex/bib/DP,bibtex/bib/GP,bibtex/bib/sparse_model,bibtex/bib/ensemble}

\end{document}